\documentclass{article}
\PassOptionsToPackage{hyphens}{url}

\usepackage{microtype}
\usepackage{graphicx}
\usepackage{booktabs}
\usepackage{multirow}
\usepackage{adjustbox}
\usepackage{makecell}
\usepackage{xcolor}
\usepackage{hyperref}

\usepackage[preprint]{icml2026}
\usepackage{amsmath,amssymb,amsthm}
\usepackage{mathtools}
\usepackage{enumitem}
\usepackage{pifont}
\usepackage[capitalize,noabbrev]{cleveref}   
\newcommand{\cmark}{\ding{51}}
\newcommand{\xmark}{\ding{55}}


\newcommand{\pis}{\pi_\theta}                 
\newcommand{\pik}{\pi^{d}_{k}}                
\newcommand{\piold}{\pi^{d}_{k,\mathrm{old}}}  
\newcommand{\piT}{\pi^{d}_{T}}                
\newcommand{\piref}{\pi_{\mathrm{ref}}}       

\newcommand{\KL}{\mathrm{KL}}
\newcommand{\M}{\mathcal{M}}
\newcommand{\D}{\mathcal{D}}

\newcommand{\method}{MRPD}
\newcommand{\methodfull}{multi-expert replay policy distillation}
\newcommand{\methodfullul}{\underline{m}ulti-expert \underline{r}eplay \underline{p}olicy \underline{d}istillation}

\icmltitlerunning{Replay, Don't Resample: Off-Policy Consolidation of Long-Horizon RL Experts}

\begin{document}

\twocolumn[
\icmltitle{Replay, Don't Resample: \\ Off-Policy Consolidation of Long-Horizon RL Experts}

\icmlsetsymbol{equal}{*\hspace{-1.65pt}}

\begin{icmlauthorlist}
\icmlauthor{Zhenrui Yue}{equal,aff1}
\icmlauthor{Gongxiaohui Chen}{equal,aff2}
\icmlauthor{Huimin Zeng}{aff1}
\icmlauthor{Honglei Zhuang}{aff3}
\icmlauthor{Jiashuo Sun}{aff1}
\icmlauthor{Zhen Han}{aff4}
\icmlauthor{Dong Wang}{aff1}
\end{icmlauthorlist}

\icmlaffiliation{aff1}{University of Illinois Urbana-Champaign}
\icmlaffiliation{aff2}{ANL}
\icmlaffiliation{aff3}{Independent Researcher}
\icmlaffiliation{aff4}{LMU}
\icmlcorrespondingauthor{}{\texttt{\{zhenrui3, dwang24\}@illinois.edu}}

\icmlkeywords{on-policy distillation, reinforcement learning, LLM agents, experience replay}

\vskip 0.3in
]

\printAffiliationsAndNotice{\icmlEqualContribution}

\begin{abstract}
Frontier post-training follows a common recipe: train one expert per domain with reinforcement learning (RL), then consolidate the experts into one model via multi-teacher on-policy distillation (MOPD). Such consolidation is expensive: the student samples in every environment again, an expert scores every sampled token, and the student inherits the drift of the final experts. Yet we argue that the experts' RL runs have already produced what consolidation needs: rollouts with their sampling probabilities and rewards, in the order the experts learned from them. We propose \methodfull{} (\method), which replays these training logs through the student in the experts' order. At each step, the policy gradient is recomputed on these scored trajectories, and the final experts' distributions are distilled at the logged prefixes. Following the experts' own trajectory reduces the training-inference mismatch and keeps the student within the clip range, so the replay is off-policy in origin but quasi-on-policy throughout consolidation. \method{} samples nothing and calls no environment or reward model: the trainer is the sole process, and consolidating the experts costs only about a seventh of the compute used to train them. On three long-horizon agentic domains (reasoning, search and SQL), \method{} yields the best student model, substantially outperforming online RL baselines, and surpassing MOPD from the same experts at a fraction of MOPD's cost. Our analysis on the same logs shows that the replayed trajectories carry most of the gain, and that both the policy gradient and the distillation term add to it. Most interestingly, the order of the logs matters more than their amount or rewards, suggesting that expert training logs are a curriculum, not a data mixture.

\end{abstract}

\section{Introduction}
\label{sec:intro}

\begin{figure}[t]
\centering
\includegraphics[width=\columnwidth, trim=5.7cm 2.9cm 5.5cm 3.2cm, clip]{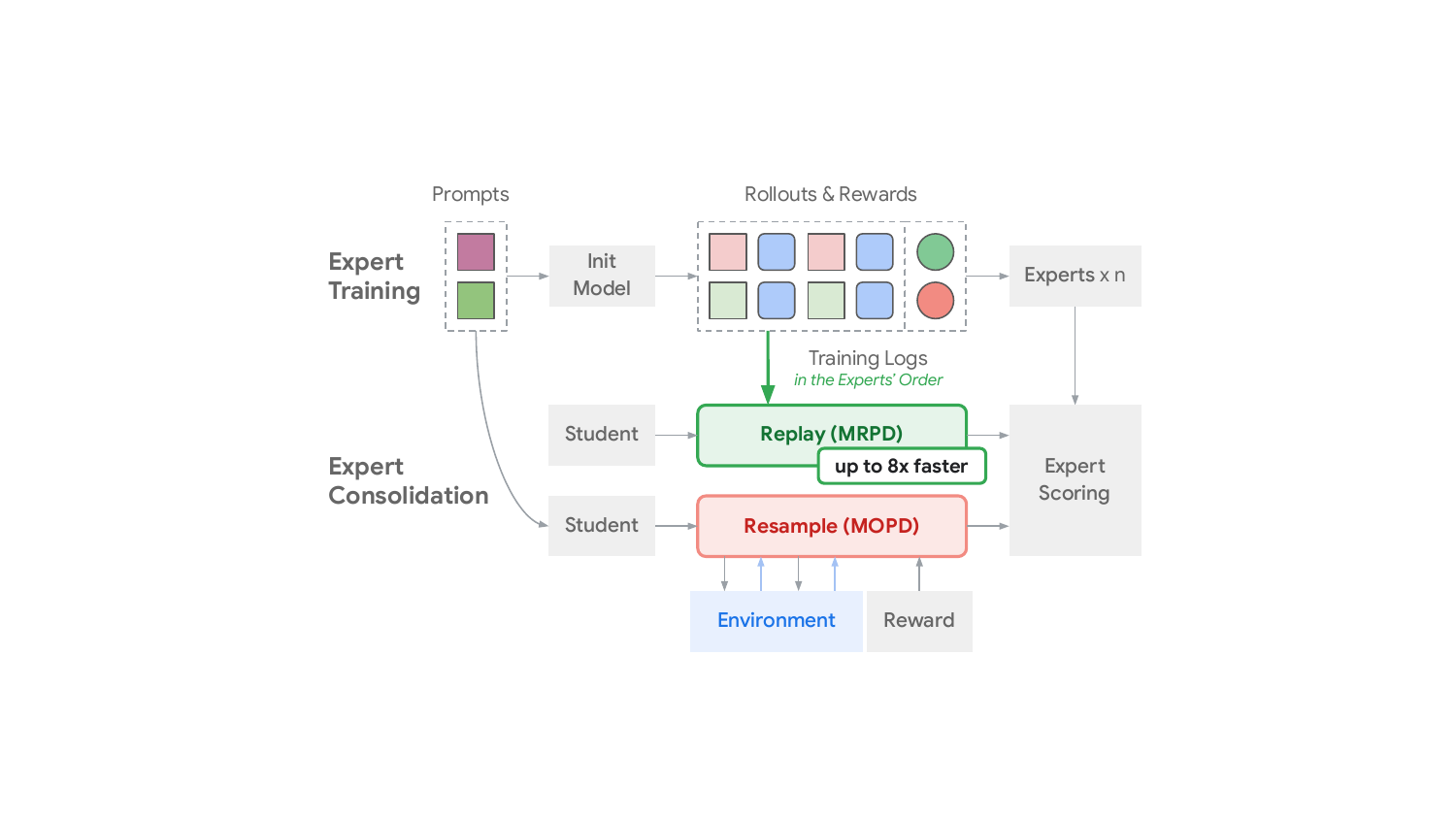}
\caption{\method{} replays instead of resampling. MOPD consolidates the experts by resampling and interacting with the environments. \method{} replays the rollouts and rewards from the experts' own training logs, in the experts' order, and can be up to 8$\times$ faster.}
\label{fig:intro}
\end{figure}
\vspace{-5pt}

Reinforcement learning (RL) with verifiable rewards has become the standard approach to fine-tuning a large language model (LLM) into a long-horizon agent that reasons with tools over multiple steps~\citep{shao2024deepseekmath,guo2025deepseek,feng2026retool,jin2025search}. However, RL on a single domain tends to improve one capability at the expense of others~\citep{yang2026local,cai2026advancing}. Frontier post-training therefore follows a common recipe: train one expert per domain from a shared supervised checkpoint, then consolidate the experts into one model via \emph{multi-teacher on-policy distillation} (MOPD), in which the student samples its own trajectories and each domain's expert scores them token by token~\citep{xiao2026mimo,team2026kimi}. This recipe has two advantages. Each expert is trained on its own domain alone, so no domain is traded off against another. And the student learns from its own samples rather than from a fixed dataset, so it is corrected on the mistakes it actually makes~\citep{agarwal2024policy,gu2023minillm}.

Nevertheless, such consolidation is expensive (\Cref{fig:intro}). First, the student samples in every environment again: the tools and sandboxes of all domains run for the whole stage, and each step waits for its longest trajectory. Second, each expert scores every token of its domain's trajectories. Third, the student is anchored to the final experts alone, so it inherits whatever drift the experts acquired in their runs. Yet we argue that the experts' RL runs have already produced what consolidation needs. Every RL step samples a batch of rollouts, scores their outcomes and records the sampling probabilities. For three experts trained for 300 steps, this amounts to about 460k trajectories in the order the experts learned from them, sampled by policies that start at the student's own initialization and end at the final experts. These training logs are normally discarded once the runs end.

Na\"{\i}vely replaying the logs raises three challenges. (1)~The trajectories are off-policy: neither the student nor the final experts sampled them, and an importance ratio to the logging policy is only usable while that policy stays close to the student. (2)~Without fresh rollouts, the usual on-policy signal, i.e., the experts' log-probability of the sampled token, has to be taken on tokens that the student did not sample, which can be less reliable. (3)~The logs are not one dataset but a sequence of batches from a sequence of different policies; whether their order carries useful learning signals decides whether they can be pooled. Our key observation is that \emph{when the student starts from the same initialization as the experts and learns on the logs in the experts' order, it stays close to the policies that sampled them.}

Building on this observation, we propose \methodfullul{} (\method), which replays the experts' logs to train the student in the experts' order (\Cref{fig:intro}). Two design choices answer the three challenges and distinguish \method{} from both MOPD and offline distillation on pooled data. First, the logs are \emph{a schedule rather than a dataset}: at step $k$, the student learns on each expert's step-$k$ batch, and the policy gradient is recomputed on these scored trajectories. Following the experts' own trajectory reduces the training-inference mismatch and keeps the student within the clip range, with at most 4\% of the tokens clipped (\Cref{sec:analysis}). Thus, the replay is off-policy in origin but \emph{quasi-on-policy} throughout consolidation. Second, the sampled-token signal is replaced by \emph{a dense target}: at every logged prefix, the final expert of the domain provides its next-token distribution, and the student minimizes the KL divergence to it.\footnote{We use ``distillation term'' and ``KL term'' interchangeably.} As a result, consolidation is fast: each step is cheap, since \method{} samples nothing and calls no environment, and every step is informative, since every token is supervised by its expert rather than by sparse per-response rewards. On three long-horizon agentic domains (reasoning, search and SQL), \method{} yields the best student model, outperforming MOPD from the same experts at a fraction of its cost. Furthermore, our analysis on the same logs shows that the replayed trajectories carry most of the gain, with the policy gradient and the dense target both adding to it, and that the order of the logs matters more than their amount or rewards.

Our contributions are summarized as follows:\footnote{Code is available at \href{https://github.com/Yueeeeeeee/MRPD}{\texttt{github.com/Yueeeeeeee/MRPD}}.}
\begin{enumerate}[leftmargin=1.5em]
\item We argue that the experts' RL runs have produced what consolidation needs, and propose \method, which replays their training logs through the student in the experts' order, with the policy gradient recomputed on the logged trajectories and the final experts' distributions distilled at the logged prefixes.
\item We show that \method{} yields the best student model on three long-horizon agentic domains: it substantially outperforms the online RL baselines, surpasses MOPD from the same experts and rivals the experts themselves. It does so at a third of MOPD's cost and about a seventh of the compute used to train the experts.
\item Through variants and ablations on the same logs, we show that the trajectories replayed in order already deliver most of the gain, with the policy gradient and the distillation target both adding to it, and that the order of the logs matters more than their amount or rewards: the order keeps the consolidation quasi-on-policy, and shuffling it causes consistent performance drops.
\end{enumerate}

\section{Related Work}
\label{sec:related}

\subsection{Model Consolidation}
Training a single model with RL across multiple domains is known to suffer from sparse signals and cross-domain interference~\citep{yang2026local,cai2026advancing}. Consequently, frontier post-training pipelines often train specialized domain experts and then consolidate them into a unified model. Model merging offers a data- and training-free approach to this consolidation, with representative methods such as model soups, task arithmetic, and TIES~\citep{wortsman2022model,ilharco2022editing,yadav2023ties} directly combining expert weights in parameter space. On the on-policy front, Cascade RL trains one model on the domains in sequence~\citep{wang2025nemotron}, while MOPD distills the experts into a single student, with each expert teaching its own domain~\citep{xiao2026mimo}. For instance, MiMo-V2-Flash rewards the student via the teacher-to-student log-ratio at each sampled token, incorporating importance sampling and an outcome reward. However, model merging often yields suboptimal performance, whereas on-policy methods incur substantial overhead on long-horizon tasks. Motivated by this, we explore off-policy consolidation by directly replaying logged expert trajectories.

\subsection{On-Policy Distillation}
On-policy distillation (OPD) samples trajectories from the student and uses the teacher's log-probability at the sampled token (or its full distribution) as a per-token signal to optimize student parameters~\citep{lu2025onpolicydistillation,agarwal2024policy,gu2023minillm}. ExOPD~\citep{yang2026learning} extrapolates beyond the teacher, while hybrid approaches combine token-level signals from an external teacher~\citep{lin2026policy,akhondzadeh2026reward} or a privileged self-teacher~\citep{yang2026self,yue2026sipo} with task rewards. For OPD-based model consolidation, MOPD methods such as Kimi K3~\citep{team2026kimi} distill multiple domain experts via sampled-token KL, whereas DeepSeek-V4~\citep{xu2026deepseek} leverages full-vocabulary KL for improved learning dynamics. Crucially, all of these approaches sample from the student policy and query the teacher throughout training. In contrast, \method{} keeps the teacher's next-token distribution as the target but does so without online rollouts: the distribution is solely evaluated on logged trajectories.

\subsection{Learning from Off-Policy Trajectories}
Sequence-level distillation trains the student on samples from the teacher model~\citep{kim2016sequence,guo2025deepseek}. Rejection-sampling fine-tuning, RAFT, STaR and ReST train on the model's own filtered samples~\citep{yuan2023scaling,xiong2025minimalist,zelikman2022star,singh2023beyond}, and LUFFY~\citep{yan2026learning} mixes off-policy expert traces into on-policy RL. \citet{wang2026building} distill several RL experts off-policy from their samples and then refine on-policy. In all of these frameworks, the trajectories are generated for the purpose of training. Experience replay~\citep{lin1992self,mnih2015human} and offline RL~\citep{levine2020offline} instead learn from trajectories that were already collected. Concurrent with our work, REGEN~\citep{chen2026regen} brings this idea to consolidation: it recycles the rollouts of single-domain GRPO experts instead of sampling new ones. It pools the rollouts of all steps into one dataset, filters it, and trains on it with an offline advantage-weighted likelihood loss. \method{} starts from a different premise: the experts' logs are not a dataset to be pooled but a curriculum to be followed. It replays the rollouts in the order the experts generated them, which keeps the student close to the policies that sampled them, and it distills the final experts' next-token distributions at the logged prefixes, a signal that offline RL on the rewards alone does not have. 
\section{Methodology}
\label{sec:method}

\begin{figure*}[t]
\centering
\includegraphics[width=0.95\textwidth, trim=1cm 2.8cm 0.4cm 2.9cm, clip]{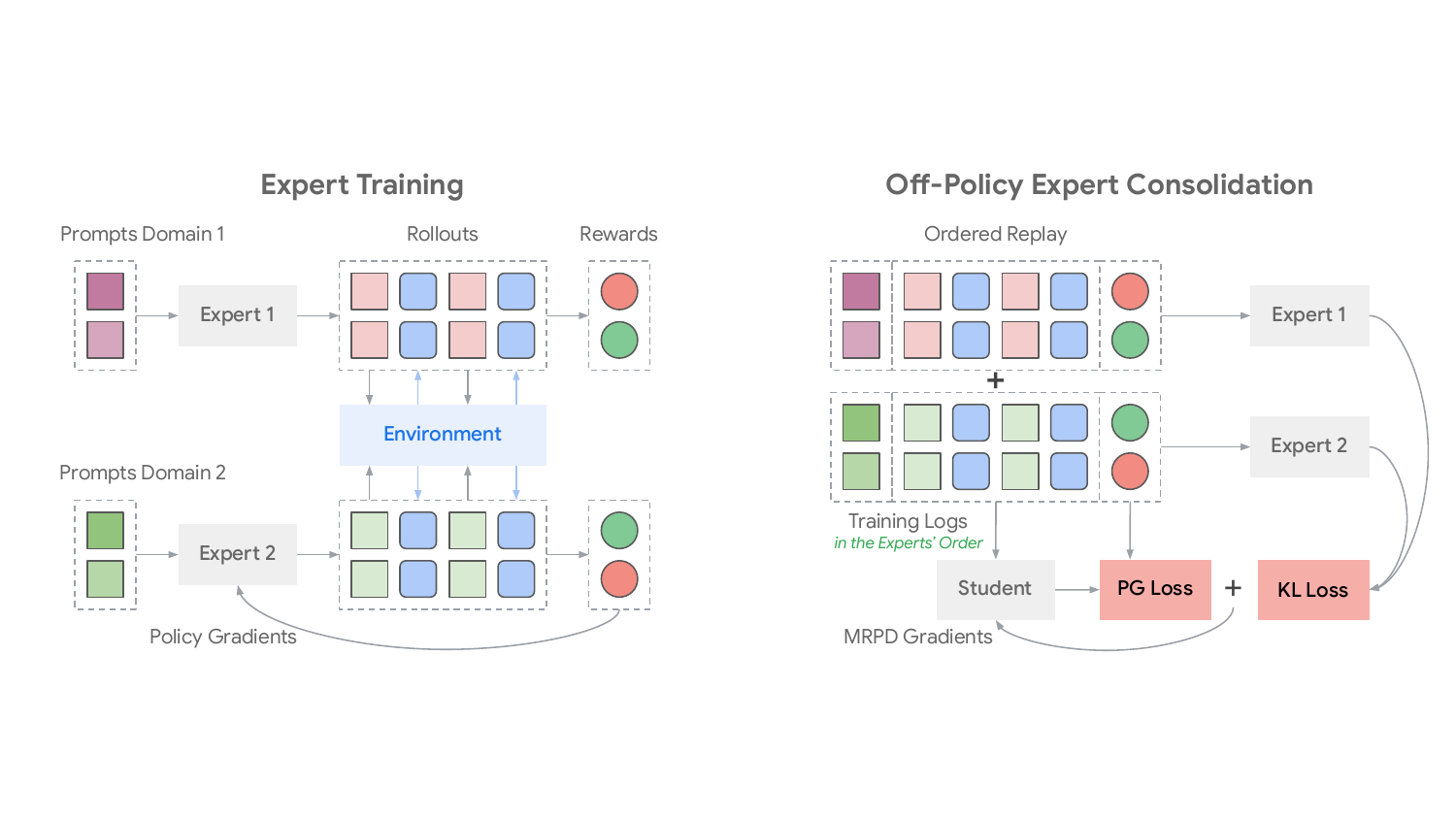}
\caption{Overview of expert RL training and off-policy expert consolidation with \method. Left: each expert is trained by RL in its own environment; every step's prompts, rollouts, log-probabilities and rewards are logged (light cells: the model's tokens; blue cells: tool outputs, excluded from the loss). Right: \method{} replays the logs in the experts' order, with the policy gradient recomputed on the logged trajectories and the final experts' distributions distilled at the logged prefixes; no environment, no sampling.}
\label{fig:method}
\end{figure*}

\subsection{Setup}
\label{sec:setup}
We consider $D$ domains, each with an environment, a tool and a binary reward $R\in\{-1,+1\}$. A shared supervised checkpoint $\piref$ initializes every model. For each domain $d$, an expert is trained by RL from $\piref$ for $K$ steps, and $\piT$ denotes the final expert ($T$ for teacher). At each step $k$, the expert's run samples a batch of trajectories from its sampling policy $\piold$, executes the tools and scores the outcomes, and then updates the expert $\pik$ on this batch. For every sample, it stores the tokens $(x, y)$, the loss mask $\M$ of the model-generated tokens (tool outputs are excluded), the reward $R$, and the sampling log-probabilities $\log\piold(y_t\mid x, y_{<t})$. We call these records the expert's \emph{training log} $\D^d=\{\D^d_k\}_{k=1}^{K}$, where $\D^d_k$ is the batch of step $k$ (\Cref{fig:method}); the experts' logs together serve as the student's replay buffer, and we use the two terms interchangeably. The goal is to train a single student $\pis$ from $\piref$ to match each expert on its respective domain, given only the experts and their logs with no further environment interaction.

\subsection{Training-Log Replay}
\label{sec:replay}
\method{} trains the student for $K$ steps. At each step $k \in \{1, \dots, K\}$, it takes one optimizer step on the aggregated batch $\D_k = \bigcup_{d=1}^D\D^d_k$ collected from the domain experts' $k$-th training step. The student thus processes expert trajectories in chronological order: early on, from policies close to $\piref$, and later, from policies approaching the final experts. In doing so, it consumes the exact sequence of experience that produced the final experts. Whenever an objective involves an importance ratio, the ratio is evaluated against the corresponding logging policy:
\begin{equation}
  r_t \;=\; \frac{\pis(y_t\mid x,y_{<t})}{\piold(y_t\mid x,y_{<t})}.
  \label{eq:ratio}
\end{equation}
This is the importance ratio of the expert's own clipped update, $\pik/\piold$, with the student $\pis$ in place of the expert $\pik$. Crucially, replaying the experts' trajectories in order keeps the student close to $\piold$: the training-inference mismatch stays largely within the clip range throughout training (\Cref{sec:analysis}). Thus, while the replay is off-policy in origin, with no student-generated samples, it remains quasi-on-policy throughout consolidation. Every loss below takes the batch average of a token-level objective averaged over $\M$, differing only in the choice of per-token term.

\subsection{The \method{} Objective}
\label{sec:losses}
\method{} combines two per-token terms: the policy gradient, recomputed on the logged trajectories and rewards (the PG term), and the distillation of the final domain expert at the logged prefixes (the KL/distillation term).

\paragraph{Policy-Gradient Term.}
As in GRPO~\citep{shao2024deepseekmath}, the advantage is the group-normalized reward of a prompt's $g$ samples, $\hat A=\frac{R-\bar R}{\sigma_R+\epsilon}$, and the per-token term is the clipped surrogate
\begin{equation*}
  -\min\big(r_t\hat A,\ \bar r_t\hat A\big),\qquad
  \bar r_t=\mathrm{clip}(r_t,1-\epsilon_{\mathrm{lo}},1+\epsilon_{\mathrm{hi}}),
\end{equation*}
with $r_t$ from \Cref{eq:ratio} and no KL penalty. Prompts whose samples all succeed or all fail yield no gradient.

\paragraph{Distillation Term.}
At every logged prefix $s_t=(x,y_{<t})$ with $t\in\M$, the distillation term pulls the student toward the final expert of the trajectory's domain:
\begin{equation}
  \ell_{\mathrm{KL}}(s_t)\;=\;\beta\,\KL\!\big(\pis(\cdot\mid s_t)\,\big\|\,
  \piT(\cdot\mid s_t)\big).
  \label{eq:kl}
\end{equation}
The KL can be evaluated at two levels. At the \emph{token level}, it is estimated from the logged token alone, $\log\pis(y_t\mid s_t)-\log\piT(y_t\mid s_t)$, as in MOPD's per-token reward~\citep{xiao2026mimo,lu2025onpolicydistillation}; the estimate enters the
advantage with a negative sign and passes through the importance ratio and the clip of the PG term. At the \emph{vocabulary level}, it is computed in closed form from the two next-token distributions and directly minimized without clipping. We call the two the sampled-token and the dense KL. By default, \method{} uses the dense KL, computed over the top-$k$ tokens and a tail atom with the remaining mass, with $k=256$ and $\beta=1$ unless stated otherwise; more details are reported in \Cref{app:impl}.

\paragraph{Objective.}
Together, \method{} at step $k$ maximizes
\begin{equation}
\begin{aligned}
&\mathcal{J}_{\method}(\theta)\;=\;\mathbb{E}_{(x,\{y_i\}_{i=1}^{g})\sim\D_k}\\
&\quad\bigg[\frac{1}{g}\sum_{i=1}^{g}\frac{1}{|\M_i|}\sum_{t\in\M_i}\Big(\min\big(r_{i,t}\hat A_i,\ \bar r_{i,t}\hat A_i\big)\\
&\qquad\qquad\qquad-\beta\,\KL\big(\pis(\cdot\mid s_{i,t})\,\big\|\,\piT(\cdot\mid s_{i,t})\big)\Big)\bigg],
\end{aligned}
\label{eq:objective}
\end{equation}
where a prompt $x$ and its group of $g$ logged responses are from the batch $\D_k$ of the replay buffer rather than from $\pis$; they were sampled by $\piold$ when the log was written, which is why the ratio follows \Cref{eq:ratio}. Here $s_{i,t}=(x,y_{i,<t})$, $\M_i$ is the loss mask of response $i$ and $\hat A_i$ its advantage. The first term applies the experts' training objective to the student on the logged data; the second pulls the student toward the final expert at every token position.

\subsection{\method{} Variants and Cost}
\label{sec:onlogs}
\label{sec:cost}
At every step, \method{} runs only the update and one forward pass of each expert on its domain's logged batch, skipping the student generation and tool calls of MOPD. The variants below keep the ordered replay of \Cref{sec:replay} and change only the per-token loss and what a step computes (measured costs are reported in \Cref{sec:results}):
(1)~\emph{Sampled-token distillation} replaces the dense KL by the token-level estimate
in \Cref{eq:kl}. The estimate gets noisier as the training-inference mismatch grows, and the clip truncates it where the mismatch is large (\Cref{sec:analysis}). It also needs the same forward pass of the experts but without storing their top-$k$.
(2)~\emph{No distillation} removes the KL term and leaves the PG term: the experts' training objective on their logged data. Its only signal is the logged reward, one scalar per trajectory. No expert is queried, so a step is the update alone.
(3)~\emph{No policy gradient} zeroes the rewards and leaves the distillation term unchanged, sampled-token or dense: off-policy distillation of the final experts at their logged prefixes. The PG term shares the policy update's forward pass, so removing it saves nothing.
(4)~\emph{Neither term} leaves the negative log-likelihood of the logged token as the per-token
term,
\begin{equation}
  \ell_{\mathrm{SFT}}(s_t)\;=\;-\log\pis(y_t\mid s_t),
  \label{eq:sft}
\end{equation}
on every sample (SFT) or on the samples with $R=1$ only. The former is the replay counterpart of supervised fine-tuning on expert samples. The latter closely resembles an offline version of RAFT~\citep{xiong2025minimalist}, which trains on the positively rewarded samples of the model's own rollouts, and we refer to it as offline RAFT. Both maximize the likelihood of the logged responses with no importance ratio: SFT imitates every response with the failed ones included, and offline RAFT uses the reward only as a filter.
(5)~\emph{Forward KL}, $\KL(\piT\,\|\,\pis)$, replaces the reverse KL in the dense term with and without the policy gradient. The reverse KL penalizes student mass where the expert has little, and the forward KL penalizes missing expert mass; both share the same top-$k$ atoms and computation cost.

\section{Experiments}
\label{sec:exp}

\begin{figure}[t]
\centering
\includegraphics[width=\columnwidth, trim=0.5cm 0.5cm 0.5cm 0, clip]{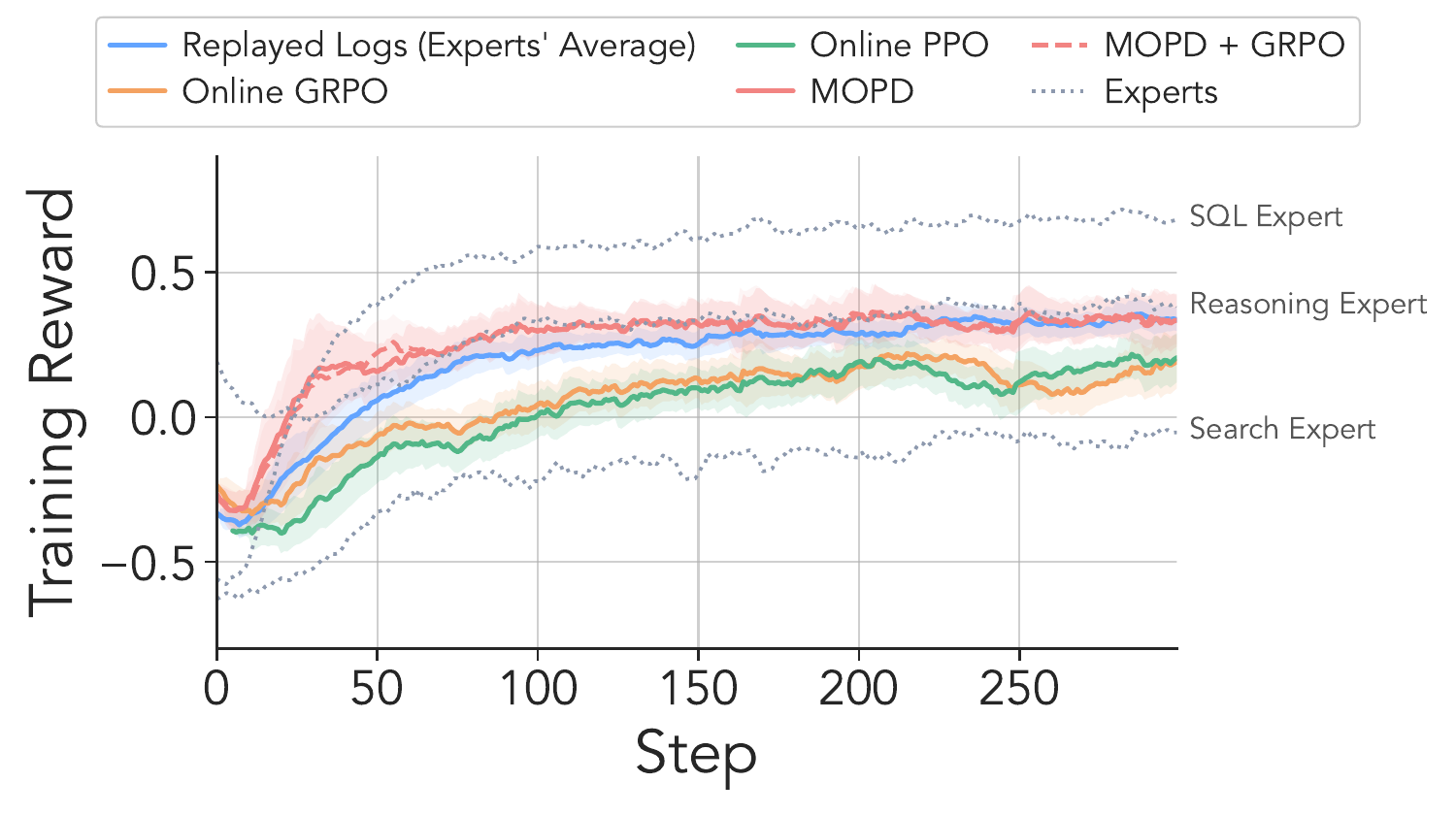}
\caption{The mean $\pm1$ training rewards of a step's rollouts: the three experts (dotted) and their average, which is the reward of the batch \method{} replays at that step; online GRPO, online PPO, MOPD and MOPD + GRPO on their own rollouts (576 per step over the three domains). MOPD does not train on the reward.}
\label{fig:reward}
\end{figure}

\subsection{Experiment Settings}
\label{sec:settings}

\paragraph{Datasets \& Models.}
We experiment on three agentic domains. \emph{Reasoning} is competition mathematics with a Python sandbox in the ReTool setting~\citep{feng2026retool}, with up to 16 code turns and 16k-token responses. \emph{Search} is open-domain QA using an E5 retriever over Wikipedia in the Search-R1 setting~\citep{jin2025search}, with up to 8 search turns. \emph{SQL} is Spider and BIRD~\citep{yu2018spider,li2023can} with a read-only SQLite tool and up to 8 turns, scored by execution. The training prompts are DAPO-Math-17k for reasoning, NQ and HotpotQA for search, and the Spider training set for SQL~\citep{yu2026dapo,jin2025search,yu2018spider}. All models start from $\piref$: Qwen3-4B~\citep{yang2025qwen3} after a supervised cold start on ReTool's SFT data. The experts are single-domain GRPO runs~\citep{shao2024deepseekmath} from $\piref$ with 64 prompts $\times$ 8 samples per step, learning rate $10^{-6}$, clip $(0.2, 0.28)$ and no KL penalty, trained asynchronously for $K=300$ steps. The replayed batch is $3\times64\times8=1536$ samples per step, and the student takes 300 steps with the experts' hyperparameters.

\paragraph{Baselines.}
We compare with three groups of baselines. The \emph{training-free} baselines are $\piref$ and the merged experts: the uniform soup~\citep{wortsman2022model} averages the experts' deltas from $\piref$, and TIES~\citep{yadav2023ties} merges them (density 0.2, $\lambda=1$). The \emph{online RL} baselines train $\piref$ on the three environments at once, with 24 prompts $\times$ 8 samples per domain and step (576 in total to keep online sampling affordable) for 300 steps: RAFT~\citep{xiong2025minimalist} trains on the correct samples of the student's own on-policy rollouts; GRPO~\citep{shao2024deepseekmath}; and PPO~\citep{schulman2017proximal} (see details in \Cref{app:impl}). The \emph{on-policy distillation} baselines sample in the same way but distill from the experts: MOPD, in which each domain's expert scores the student's trajectories at the sampled token, and MOPD + GRPO, which adds the task reward as in~\citet{xiao2026mimo}. The \emph{domain experts}, i.e., the GRPO experts and single-domain PPO runs, serve as single-domain references, each evaluated on its own domain.

\begin{figure}[t]
\centering
\includegraphics[width=\columnwidth]{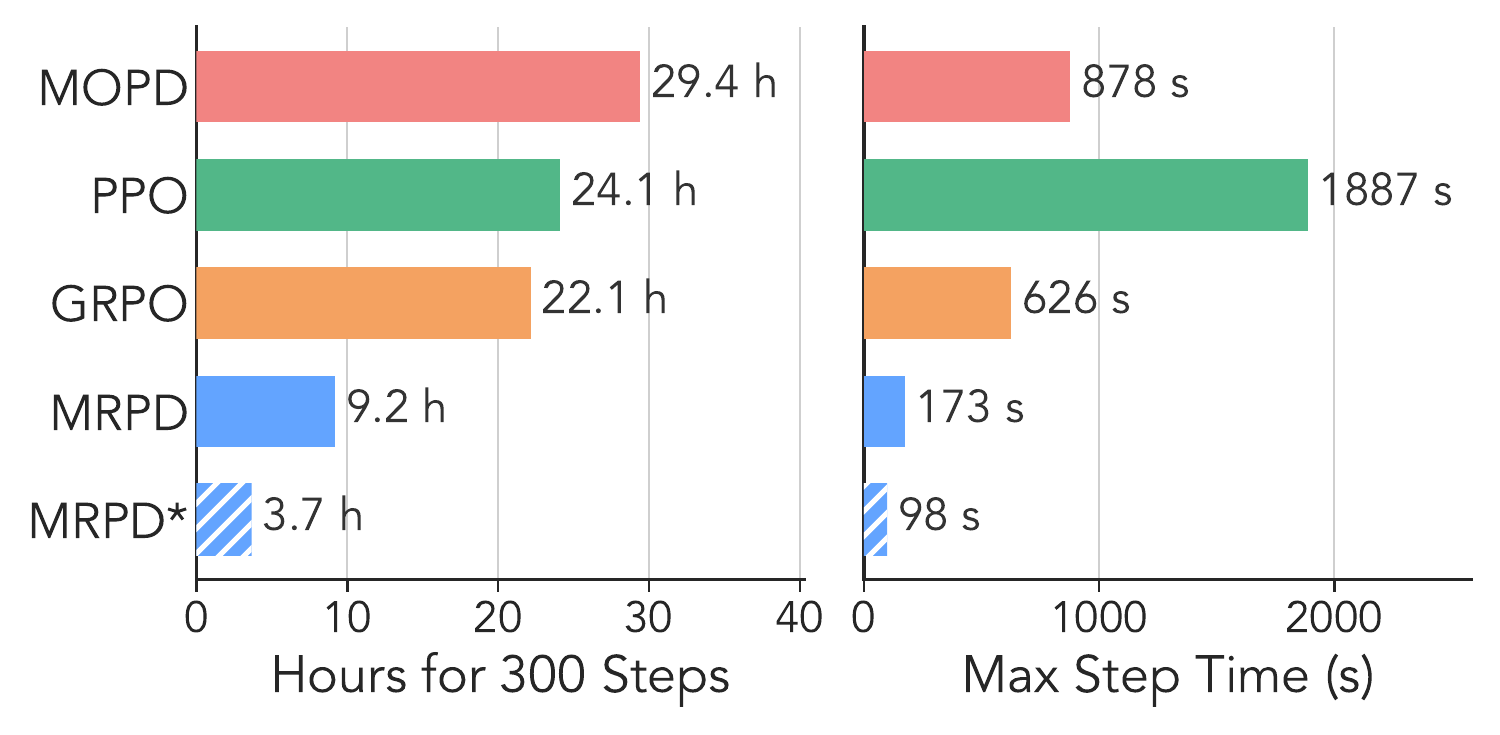}
\caption{Cost on eight H100s: the wall-clock time of the full training (left) and the longest single step (right). MOPD and the online RL baselines (PPO, GRPO) sample rollouts at every step; \method{} replays the logs with the full 1536 samples per step, and \method$^*$ is data-matched to the on-policy methods (576 samples per step). MOPD is timed on the run with task reward (MOPD + GRPO), which samples and runs the experts exactly as MOPD does. PPO's longest step is a single slow rollout.}
\label{fig:cost}
\end{figure}

\begin{table*}[t]
\centering
\caption{Main results after 300 steps: five of the six reasoning sets (AIME 2024, AIME 2025 and AMC 2023 at 32 samples per problem, OlympiadBench and MATH-500 at 4), five of the seven search sets (exact match, one sample) and the three SQL sets (execution accuracy, 4 samples); the other sets are in \Cref{app:sets}. Average is the mean over \emph{all} evaluation sets of the domain, and Overall denotes the mean of the three domain averages. Bold: the best trained student; underline: the second best. The domain experts are single-domain RL references and are not ranked, with the GRPO experts being the teachers and providing training logs.}
\label{tab:main}
\setlength{\tabcolsep}{4.5pt}
\adjustbox{max width=\linewidth}{%
\begin{tabular}{@{}lccccccccccc@{}}
\toprule
\multirow{3}{*}{\textbf{Dataset}} & \multicolumn{3}{c}{\textbf{Training-Free}} & \multicolumn{5}{c}{\textbf{Online RL / Distillation}} & \multirow{3}{*}{\makecell{\textbf{\method}\\\textbf{(Ours)}}} & \multicolumn{2}{c}{\textbf{Domain Experts}} \\
\cmidrule(lr){2-4}\cmidrule(lr){5-9}\cmidrule(lr){11-12}
 & \makecell{SFT\\Init.} & \makecell{Uniform\\Soup} & \makecell{TIES\\Merging} & \makecell{RAFT} & \makecell{GRPO} & \makecell{PPO} & \makecell{MOPD} & \makecell{MOPD +\\GRPO} &  & \makecell{PPO} & \makecell{GRPO} \\
\midrule
\multicolumn{12}{c}{Reasoning (Average over all six sets)} \\ \midrule
AIME 2024 & 26.9 & 39.0 & 47.0 & 41.0 & 40.3 & 39.5 & 48.1 & \underline{48.6} & \textbf{50.4} & 43.8 & 53.8 \\
AIME 2025 & 20.3 & 31.4 & 38.0 & 32.4 & 32.8 & 31.9 & \textbf{42.4} & \underline{40.7} & \underline{40.7} & 44.6 & 42.5 \\
AMC 2023 & 66.6 & 78.8 & 84.2 & 78.9 & 76.0 & 77.7 & \textbf{87.8} & \underline{87.6} & 87.5 & 89.1 & 90.3 \\
OlympiadBench & 50.4 & 59.4 & 62.2 & 60.1 & 47.1 & 54.1 & \textbf{67.3} & 66.1 & \underline{67.0} & 66.5 & 66.3 \\
MATH-500 & 81.3 & 87.8 & 87.9 & 88.1 & 74.5 & 82.0 & 92.3 & \textbf{93.0} & \underline{92.8} & 93.0 & 93.4 \\
Average & 47.3 & 56.6 & 60.0 & 57.2 & 50.3 & 53.1 & \textbf{63.8} & 63.4 & \underline{63.7} & 63.6 & 65.2 \\
\midrule
\multicolumn{12}{c}{Search (Average over all seven sets)} \\ \midrule
NQ & 17.0 & 23.8 & \textbf{38.6} & 29.3 & 35.7 & 32.7 & 37.0 & \underline{38.4} & 38.3 & 36.3 & 39.6 \\
TriviaQA & 44.8 & 52.4 & \textbf{66.0} & 59.6 & 64.6 & 62.4 & 62.6 & \underline{64.6} & 64.4 & 62.4 & 65.1 \\
HotpotQA & 19.2 & 25.8 & 40.6 & 31.7 & 37.9 & 36.0 & \underline{40.6} & 40.0 & \textbf{42.1} & 40.2 & 42.3 \\
2WikiMQA & 26.2 & 31.6 & \underline{44.4} & 34.9 & 41.0 & 39.8 & 42.3 & 43.5 & \textbf{44.5} & 41.4 & 44.2 \\
Bamboogle & 24.8 & 35.2 & \textbf{48.0} & 40.0 & 44.0 & 44.8 & 44.8 & 46.4 & \underline{47.2} & 42.4 & 50.4 \\
Average & 22.4 & 29.4 & \textbf{42.6} & 34.1 & 39.7 & 38.3 & 40.6 & 41.4 & \underline{42.0} & 39.8 & 43.0 \\
\midrule
\multicolumn{12}{c}{SQL (Average over the three sets)} \\ \midrule
Spider dev & 17.3 & 61.7 & 76.1 & 72.3 & 76.0 & 75.1 & \underline{80.1} & 79.6 & \textbf{80.7} & 79.7 & 80.3 \\
Spider test & 20.0 & 63.9 & 77.8 & 73.3 & 77.2 & 76.6 & \textbf{80.1} & \underline{79.9} & 79.8 & 79.1 & 80.0 \\
BIRD dev & 7.5 & 31.8 & 49.1 & 43.9 & 49.1 & 47.3 & 49.4 & \underline{49.6} & \textbf{49.9} & 47.8 & 48.6 \\
Average & 14.9 & 52.5 & 67.7 & 63.2 & 67.4 & 66.3 & \underline{69.9} & 69.7 & \textbf{70.1} & 68.8 & 69.6 \\
\midrule
\textbf{Overall} & 28.2 & 46.1 & 56.8 & 51.5 & 52.5 & 52.6 & 58.1 & \underline{58.2} & \textbf{58.6} & 57.4 & 59.3 \\
\bottomrule
\end{tabular}
}
\end{table*}

\paragraph{Evaluation.}
We evaluate on sixteen held-out sets after 300 steps: AIME 2024 \& 2025 and AMC 2023 by accuracy averaged over 32 samples per problem, and Minerva~\citep{lewkowycz2022solving}, OlympiadBench~\citep{he2024olympiadbench} and MATH-500~\citep{hendrycks2021measuring,lightman2023let} over 4; NQ~\citep{kwiatkowski2019natural}, TriviaQA~\citep{joshi2017triviaqa}, PopQA~\citep{mallen2023not}, HotpotQA~\citep{yang2018hotpotqa}, 2WikiMultiHopQA (2WikiMQA)~\citep{ho2020constructing}, MuSiQue~\citep{trivedi2022musique} and Bamboogle~\citep{press2023measuring} by exact match; Spider dev and Spider test~\citep{yu2018spider} and BIRD dev~\citep{li2023can} by execution accuracy at 4 samples (evaluation details reported in \Cref{app:eval}).

\subsection{Main Results}
\label{sec:results}

\paragraph{Training Reward.}
\Cref{fig:reward} shows the reward every run trains on: the experts', whose average is the reward of the batch \method{} replays at that step, and the online baselines' rewards on their own rollouts. All curves start from the reward of $\piref$, about $-0.4$. The two MOPD runs rise fastest and pass 0.2 within about 30 steps, since the final experts provide token-level learning signals from the first step; the replayed logs record the experts while they were still learning and pass 0.2 only after about 75 steps. Over the last 50 steps the three are level at about 0.34: MOPD, which does not train on the reward, ends level with MOPD + GRPO and with \method{} (i.e., replayed logs). Online GRPO and PPO climb slowest, pass 0.2 only around step 200 and end at 0.14 and 0.19, well below the experts' average. The experts themselves differ by domain: the SQL expert starts below $-0.5$ and ends at 0.69; the reasoning expert, in the one domain $\piref$ was fine-tuned on, starts near 0 and ends at 0.38; the search expert climbs slowly throughout, from $-0.6$ to $-0.07$. The online runs' reward is on their own rollouts of the training prompts and the replay's on the experts', so it does not rank the methods; the held-out sets of \Cref{tab:main} do.

\paragraph{Training Cost.}
\Cref{fig:cost} compares the cost of the 300 steps on eight H100s. A step of MOPD + GRPO samples, calls the tools and runs the experts exactly as MOPD does. This takes 350s on average, of which 340s is generation: every step waits for its longest trajectory, up to 16k tokens. A step of online GRPO takes 270s and of online PPO 290s. A step of \method{} takes 110s: the update 79s, the experts' forward 24s and 6s reading the next step's log. Over 300 steps, MOPD + GRPO takes 29 hours, online GRPO 22, online PPO 24 and \method{} 9. The experts took 30, 32 and 4.5 hours for the 300 steps we replay, so consolidating them costs 14\% of the compute used to train them, about a seventh. In the data-matched setting (576 samples per step), a step of \method{} reduces to 44s and the full training only takes 3.7 hours: an eighth of MOPD + GRPO's and a sixth of online GRPO's and PPO's. No step of \method{} takes longer than 173s, while the longest step of each online run, a slow rollout, takes 626 to 1887s. \Cref{app:steptime} shows the step times over training with more detailed explanations.

\begin{table*}[t]
\centering
\caption{Variants of \method{} on the same logs: the policy-gradient (PG) term crossed with the distillation (KL) term to the final experts, after 100, 200 and 300 steps. The KL term is the sampled-token estimate or the dense $(k+1)$-atom KL, in the reverse or the forward direction; without the PG term, the rewards are zeroed. The losses without the KL term are in \Cref{tab:order}. Domain averages over all sets and their mean (Overall). We mark the best of each column in bold and underline the second best.}
\label{tab:variants}
\setlength{\tabcolsep}{3.5pt}
\adjustbox{max width=\textwidth}{%
\begin{tabular}{@{}clcccccccccccc@{}}
\toprule
 & & \multicolumn{4}{c}{\textbf{After 100 Steps}} & \multicolumn{4}{c}{\textbf{After 200 Steps}} & \multicolumn{4}{c}{\textbf{After 300 Steps}} \\
\cmidrule(lr){3-6}\cmidrule(lr){7-10}\cmidrule(lr){11-14}
\textbf{PG} & \textbf{KL Term} & \textbf{Reasoning} & \textbf{Search} & \textbf{SQL} & \textbf{Overall} & \textbf{Reasoning} & \textbf{Search} & \textbf{SQL} & \textbf{Overall} & \textbf{Reasoning} & \textbf{Search} & \textbf{SQL} & \textbf{Overall} \\
\midrule
\xmark & sampled-token & 57.6 & 40.6 & 69.0 & 55.7 & 58.3 & 39.9 & 69.1 & 55.7 & 62.9 & 40.8 & 69.5 & 57.7 \\
\cmark & sampled-token & 55.1 & 40.2 & 67.5 & 54.3 & 60.7 & 40.7 & 69.0 & 56.8 & \underline{63.5} & 41.5 & 69.7 & 58.2 \\
\addlinespace[2pt]
\xmark & dense, reverse & \textbf{63.0} & \underline{40.9} & \textbf{69.8} & \textbf{57.9} & \textbf{63.4} & \textbf{41.8} & \textbf{69.9} & \textbf{58.4} & \underline{63.5} & 41.6 & 69.5 & 58.2 \\
\cmark & dense, reverse & 62.6 & \textbf{41.1} & 69.2 & 57.6 & \underline{63.1} & 41.4 & 69.0 & 57.8 & \textbf{63.7} & \textbf{42.0} & \textbf{70.1} & \textbf{58.6} \\
\addlinespace[2pt]
\xmark & dense, forward & \underline{62.9} & 40.8 & \underline{69.5} & \underline{57.7} & \textbf{63.4} & 41.3 & \underline{69.5} & \underline{58.1} & 63.4 & \underline{41.7} & \underline{69.9} & \underline{58.3} \\
\cmark & dense, forward & 62.0 & \underline{40.9} & 68.9 & 57.3 & 62.5 & \underline{41.5} & 69.3 & 57.7 & 62.7 & 41.4 & 69.5 & 57.8 \\
\bottomrule
\end{tabular}
}
\end{table*}

\begin{figure}[h]
\centering
\includegraphics[width=\columnwidth]{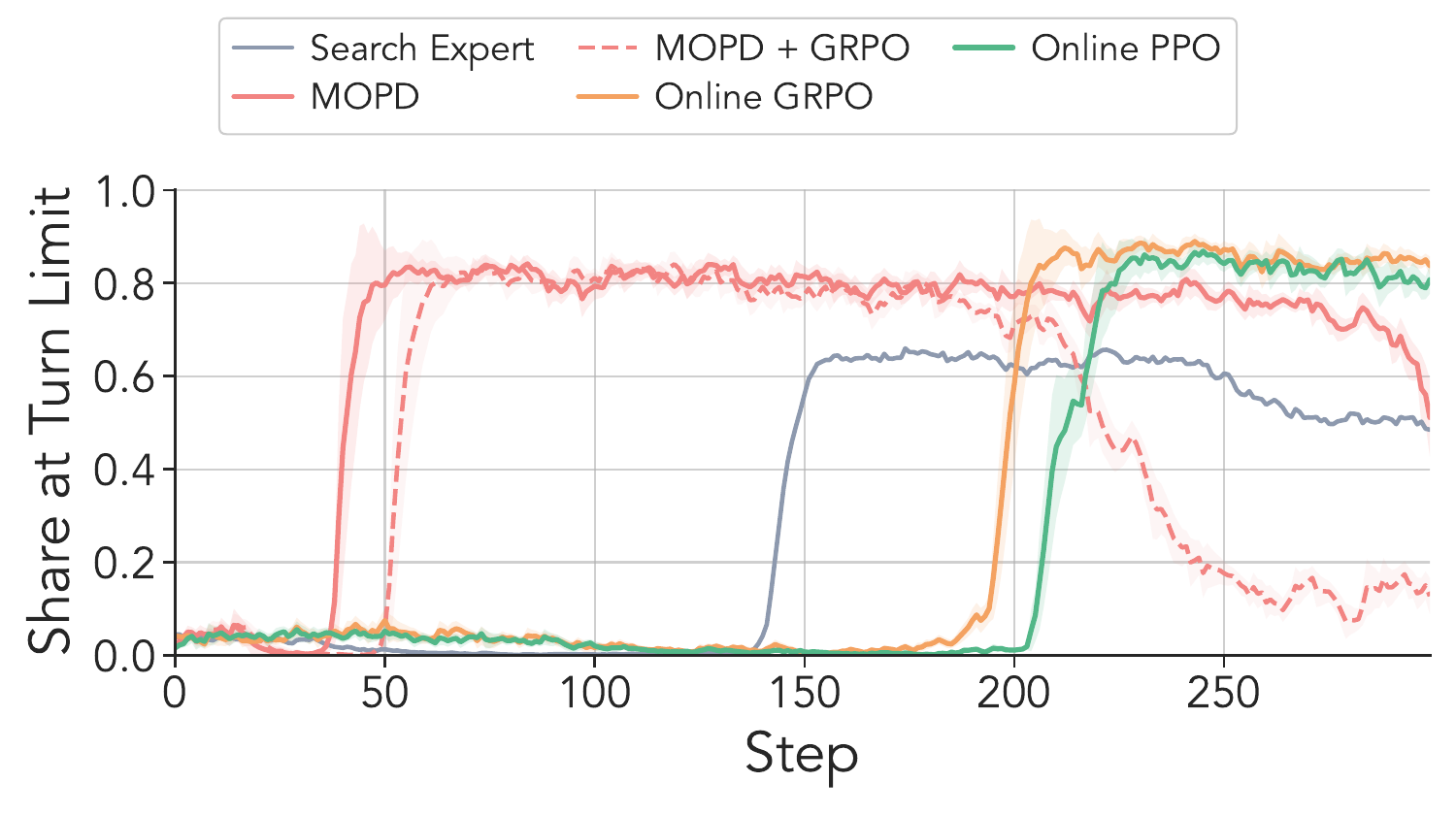}
\caption{Search rollouts cut off at the turn limit in training, as a share of the step's search trajectories: the search expert, MOPD and MOPD + GRPO (which distill from the final experts), and online GRPO and PPO. The expert reaches the limit at step 150; MOPD's student is there from step 50 and stays; online RL finds the same behavior at step 200. \method{} samples nothing and has no such curve.}
\label{fig:truncation}
\end{figure}

\paragraph{Benchmark Performance.}
\Cref{tab:main} reports the main results of \method{} and all baseline methods. Overall, \method{} yields the best student model: 58.6 on average, and the only student in the top two on every domain. Our observations: (1)~\emph{Online RL}: RAFT, GRPO and PPO reach 51.5, 52.5 and 52.6 after 300 steps, far below the single-domain GRPO and PPO experts (59.3 and 57.4); \method{} is at least six points above each. (2)~\emph{On-policy distillation}: MOPD and MOPD + GRPO from the same experts reach 58.1 and 58.2; \method{} surpasses both at a third of MOPD's cost per step (see \Cref{fig:cost}). MOPD has the best reasoning of any student (63.8) but the lowest search of the three (40.6), as its student inherits the drift of the final search expert: the expert hits its turn limit in about 60\% of its rollouts, and MOPD's student in 80\% from step 50 on (\Cref{fig:truncation}). The replay distills at the logged prefixes and does not inherit it, and the task reward of MOPD + GRPO counters it late in training (41.4 on search). At MOPD + GRPO's loss and 576 samples per step, the replay is 0.5 behind (57.7, \Cref{tab:ablation}) and reaches 58.5 with the dense KL term. (3)~\emph{Merging}: without training, the uniform soup and TIES lift $\piref$ from 28.2 to 46.1 and 56.8. The soup collapses on search and SQL (29.4 and 52.5); TIES has the best search average of any student (42.6 against 42.0 for \method) but is five points below the reasoning expert (60.0), and \method{} is above both merges overall. (4)~\emph{Experts}: \method{} is 0.7 below the GRPO experts whose logs it replays and 1.2 above the PPO experts. The GRPO experts are stronger than the PPO experts in every domain, although the PPO experts' rollouts are shorter on average in reasoning and search and comparable in SQL.

\subsection{Policy Gradient and Distillation Variants}
\label{sec:variants}

\begin{figure*}[t]
\centering
\includegraphics[width=\textwidth]{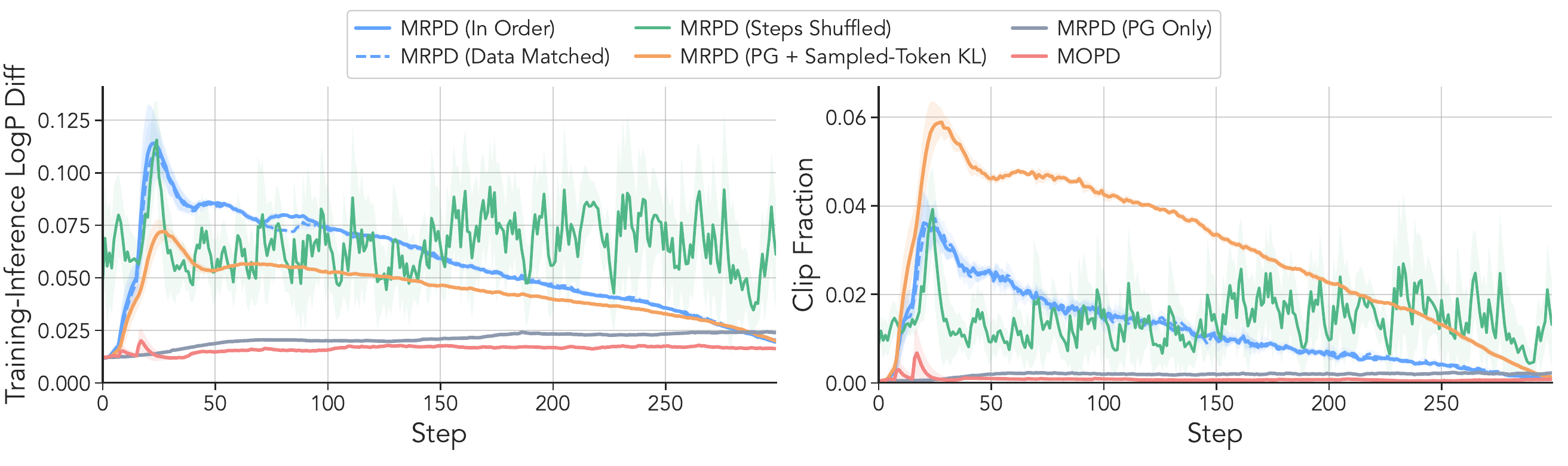}
\caption{Training-inference mismatch. Left: the mean absolute difference between the student's and the logged sampling log-probabilities on the training tokens, for the ordered replay, its data-matched variant (576 samples per step, dashed), the same replay with the sampled-token KL, with the PG term alone (offline GRPO), the shuffled replay, and MOPD (whose mismatch stems from the one-step lag in asynchronous training and training-inference engine discrepancies). Right: the clip fraction of the PG term. The ordered replay climbs, then decreases toward MOPD's level; the shuffled replay stays high on both. The sampled-token term peaks lower but clips more: it passes through the ratio, the dense term does not. The PG-only \method{} stays near MOPD's level throughout: the KL term is what opens the gap.}
\label{fig:mismatch}
\end{figure*}

\begin{table}[t]
\centering
\caption{Different replay setups without the KL term, replayed in the experts' original order or with the 300 steps shuffled. We compare negative log-likelihood on all logged responses (SFT) or only correct ones (RAFT), while using the PG term alone yields offline GRPO. On shuffled logs, \emph{PG only (offline GRPO)} is the closest counterpart to REGEN~\citep{chen2026regen} in our setting.}
\label{tab:order}
\adjustbox{max width=\columnwidth}{%
\begin{tabular}{@{}lccccc@{}}
\toprule
\textbf{Setting} & \textbf{Reasoning} & \textbf{Search} & \textbf{SQL} & \textbf{Overall} & $\boldsymbol{\Delta}$ \\
\midrule
\multicolumn{6}{@{}l}{\textit{SFT on the logs}} \\
\quad in order & 62.6 & 40.5 & 69.3 & 57.5 & -- \\
\quad steps shuffled & 57.7 & 38.3 & 65.1 & 53.7 & $-3.8$ \\
\midrule
\multicolumn{6}{@{}l}{\textit{RAFT on the logs}} \\
\quad in order & 62.8 & 41.0 & 69.1 & 57.6 & -- \\
\quad steps shuffled & 59.4 & 38.8 & 66.9 & 55.0 & $-2.6$ \\
\midrule
\multicolumn{6}{@{}l}{\textit{PG only (offline GRPO)}} \\
\quad in order & 62.6 & 41.0 & 69.6 & 57.7 & -- \\
\quad steps shuffled & 58.4 & 39.6 & 65.8 & 54.6 & $-3.1$ \\
\bottomrule
\end{tabular}
}
\end{table}

\paragraph{The Replayed Trajectories Carry Most of the Gain.}
\Cref{tab:variants,tab:order} vary the loss on the same logs. With both terms removed (\Cref{tab:order}), the objective reduces to negative log-likelihood: SFT on all logged samples, or RAFT on the correct ones. Both replay variants surpass online RL, scoring 57.5 and 57.6 respectively. The cleanest comparison is offline RAFT (\Cref{tab:order}) versus online RAFT in \Cref{tab:main}, which achieves 57.6 on the logs compared to 51.5 on the student's own rollouts. This is not a matter of sample count (\Cref{sec:analysis}); rather, at every step, the logged trajectories come from experts that are further ahead on their respective tasks than the multi-domain student. While training on expert-generated data is a classical recipe~\citep{kim2016sequence}, our key distinction is replaying these logs in order as free byproducts of expert RL, requiring no extra rollout.

\paragraph{Policy Gradient and Distillation Both Add to It.}
On top of SFT on the logs, the distillation term adds the most: alone, the sampled-token KL reaches 57.7 and the dense KL 58.2. Adding the PG term lifts them to 58.2 and 58.6, so \method{} ends 1.2 above SFT, two thirds of the way to the experts. Without the KL term, the reward adds little: RAFT's reward filter and the PG term alone reach 57.6 and 57.7. The two terms also differ over time. Similar to MOPD, the dense KL alone learns the fastest and then stalls (57.9, 58.4 and 58.2 at 100, 200 and 300 steps). With the PG term, \method{} starts slower (57.6 and 57.8 at 100 and 200 steps) but keeps improving over the last 100 steps to 58.6.

\begin{table}[t]
\centering
\caption{Ablations of the weight $\beta$ of the KL term, the amount of the logs (24 of the 64 prompts of each expert's step, i.e., 576 samples per step) and their order (the 300 steps shuffled), for the dense and the sampled-token KL, both with the PG term. $\Delta$: the change of the overall average from the default ($\beta=1$, 1536 samples per step, in order).}
\label{tab:ablation}
\small
\adjustbox{max width=\columnwidth}{%
\begin{tabular}{@{}lcccc@{}}
\toprule
\textbf{Setting} & \textbf{Reasoning} & \textbf{Search} & \textbf{SQL} & $\boldsymbol{\Delta}$ \\
\midrule
\multicolumn{5}{@{}l}{\textit{Dense KL (\method)}} \\
\quad default ($\beta=1$) & 63.7 & 42.0 & 70.1 & -- \\
\quad $\beta=3$ & 63.7 & 41.2 & 69.5 & $-0.5$ \\
\quad $\beta=0.3$ & 62.8 & 41.6 & 70.0 & $-0.5$ \\
\quad $\beta=0.1$ & 62.5 & 41.7 & 69.5 & $-0.7$ \\
\quad 576 / step & 63.6 & 41.6 & 70.1 & $-0.2$ \\
\quad steps shuffled & 63.2 & 41.4 & 69.4 & $-0.6$ \\
\midrule
\multicolumn{5}{@{}l}{\textit{Sampled-token KL}} \\
\quad default ($\beta=1$) & 63.5 & 41.5 & 69.7 & -- \\
\quad 576 / step & 62.8 & 40.8 & 69.5 & $-0.5$ \\
\quad steps shuffled & 61.4 & 41.0 & 69.3 & $-1.0$ \\
\bottomrule
\end{tabular}
}
\end{table}

\paragraph{Dense Beats Sampled-Token.}
The dense term computes the reverse KL that the sampled-token term estimates from one importance-weighted token (\Cref{sec:onlogs}). It is better by 0.5 without the PG term and by 0.4 with it. Three things separate the two terms: the sampled-token term constrains one coordinate per position, the student's log-probability of the logged token; its estimate is unbiased only for tokens sampled from the student, whereas the replayed tokens come from the logging policy; and it passes through the ratio, so a token on which the student has moved past the clip range gets no gradient (\Cref{fig:mismatch}, right). The dense term instead constrains the top-$k$ distribution in closed form without clipping, and quickly brings the student to the experts' neighborhood (\Cref{fig:domains}). The direction of the dense KL, by contrast, is a second-order choice that matters only together with the PG term. Alone, the forward and the reverse KL are level (58.3 and 58.2); with the PG term, the reverse KL gains (58.6) and the forward KL loses (57.8), the two separating over the last 100 steps. We conjecture that the direction matters late in the replay, when the student is close to the experts and its remaining errors lie in its own high-probability tokens~\citep{agarwal2024policy,gu2023minillm}: the mode-seeking reverse KL accommodates the PG term's updates, while the mass-covering forward KL resists them.

\subsection{Additional Analysis}
\label{sec:analysis}

\paragraph{Weight of the Distillation Term.}
\Cref{tab:ablation} varies the weight $\beta$ of the distillation term around 1. The dense term loses 0.5 at $\beta=3$, 0.5 at $\beta=0.3$ and 0.7 at $\beta=0.1$. With the KL term alone ($\beta=\infty$, \Cref{tab:variants}), performance also drops. Therefore, equal weights on both terms ($\beta=1$) are the best setting we tried, and a factor of three either way costs about half a point.

\begin{figure}[t]
\centering
\includegraphics[width=\columnwidth]{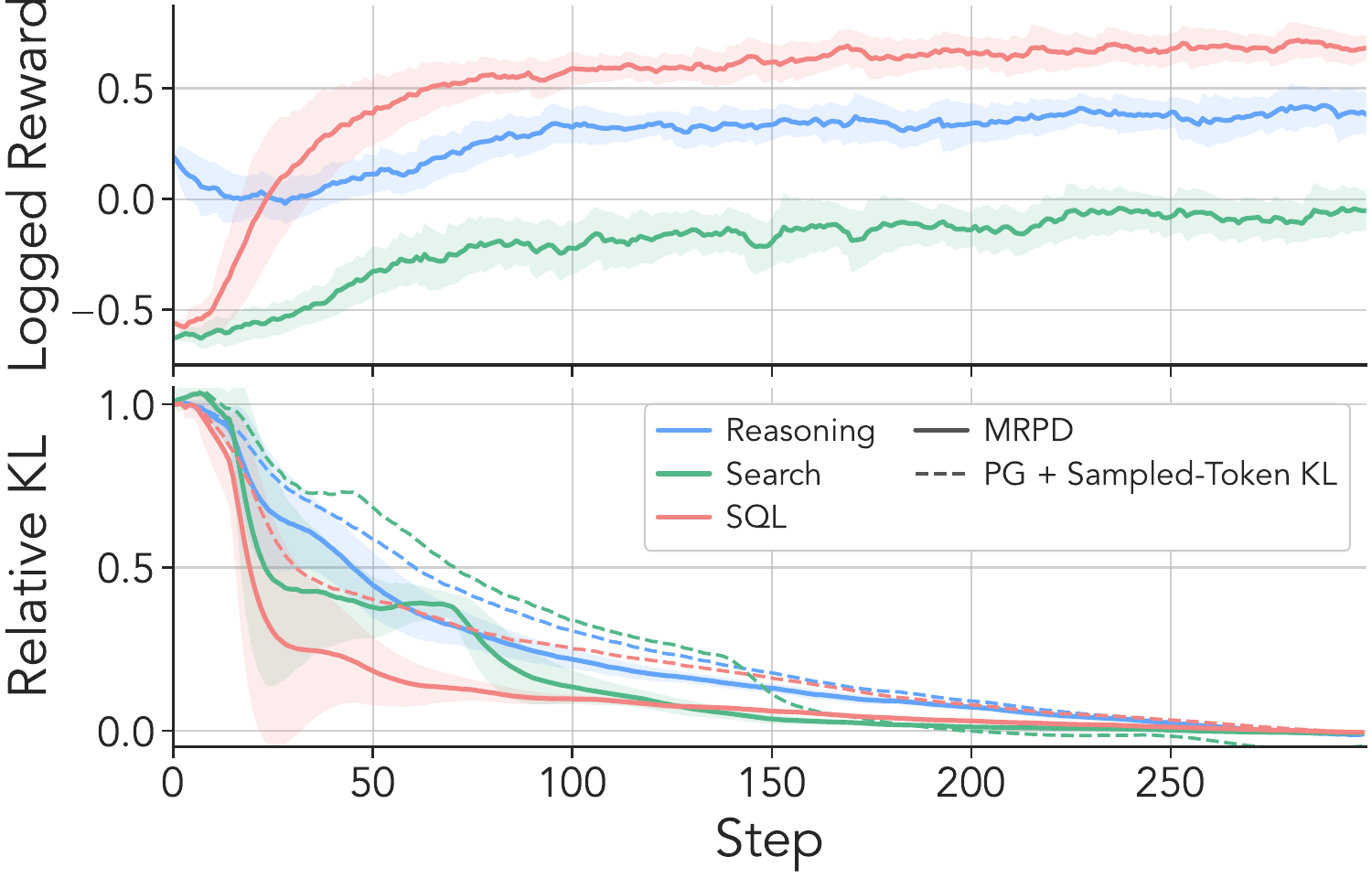}
\caption{Per-domain curves of the replay. Top: the reward of the replayed batch (the expert's own training reward at that step). Bottom: the student's KL to the domain's final expert at the replayed tokens (sampled-token estimate), relative to step 0, under the dense KL (\method, solid) and under the sampled-token KL with the PG term (dashed). The SQL expert saturates within about 100 steps. The dense term closes most of the gap within 20 steps on search and SQL, where the sampled-token term is still at a quarter after 100; the gap to the reasoning expert closes last.}
\label{fig:domains}
\end{figure}

\paragraph{Quasi-On-Policy Throughout.}
\Cref{fig:mismatch} tracks how far the student deviates from the logging policy: the mismatch (left), i.e., the mean absolute difference between the student's and the logged log-probabilities, and the fraction of tokens the PG term clips (right). The mismatch starts at 0.012 due to the inference engine's numerical difference from the trainer. With the PG term alone (offline GRPO), it never exceeds 0.026 and the clip fraction 0.003: replaying the experts' data with their own loss keeps the student close to MOPD's level (around 0.017). The KL term opens a gap by pulling the student toward the final experts. With the dense term, the student's KL to them falls by three quarters within 25 steps, fastest on search and SQL (\Cref{fig:domains}), while the logged responses of these steps still come from early experts, so the mismatch peaks around step 20, at 0.126, with 4\% of the tokens clipped; it then decreases steadily as the logs ``catch up'', to 0.019 and the clip fraction to 0.0007 at the end. With the sampled-token term the peak is lower, 0.075, but the clip fraction higher: that term passes through the ratio and is clipped with the PG term, which also slows its distillation (\Cref{fig:domains}). In other words, the dense term moves many tokens a little, mostly within the clip range, while the sampled-token term moves fewer tokens further, making their gradients more likely to be cut off. Overall, the replay is quasi-on-policy throughout: the mean log-ratio to the logging policy stays within the clip range, at most 4\% of the tokens are clipped, and the gap closes as the logs reach the final experts. Only the shuffled replay consistently stays at a high level, as discussed below.

\paragraph{Order Matters More than Amount or Rewards.}
Shuffling with the samples unchanged costs 0.6 with the dense term and 1.0 with the sampled-token term (\Cref{tab:ablation}). Shuffling does not slow the student down but caps it: after 200 steps the shuffled runs are level with or ahead of the ordered ones (58.1 against 57.8 with the dense KL, 56.9 against 56.8 with the sampled-token KL), and then stall ($-0.1$ and $+0.3$ over the last 100 steps), while the ordered runs gain 0.8 and 1.4. In the shuffled replay the student meets batches from every stage of the experts' training, so its mismatch and clip fraction stay high (\Cref{fig:mismatch}). The amount of data and the rewards matter less: keeping 24 of the 64 prompts of each expert's step (576 samples per step) costs 0.2 with the dense term and 0.5 with the sampled-token term, and zeroing the logged rewards (no PG term, \Cref{tab:variants}) 0.4 and 0.5. At 576 samples per step, the ordered replay still reaches 58.5 and 57.7, matching MOPD and above the full logs shuffled (58.0 and 57.2). Without the KL term the order matters most: shuffling costs SFT, RAFT and the PG term alone 3.8, 2.6 and 3.1 points (\Cref{tab:order}), as these losses fit the mixture of all logged policies instead of ending on the experts' last steps. The KL term anchors every step to the final experts and makes the replay more robust to the order.

\paragraph{Reasoning Is the Hardest Domain.}
\method{} trails its experts by 1.5 on reasoning and 1.0 on search, and leads by 0.5 on SQL (\Cref{tab:main}). SQL is saturated: every variant in \Cref{tab:variants} is within 0.6 points of the SQL expert. Search is where the experts' distribution helps most (+1.5 over SFT on the logs), and half of the remaining gap is Bamboogle (47.2 against 50.4). Reasoning keeps the largest gap, on the three competition sets (AIME 2024 50.4 against 53.8, AIME 2025 40.7 against 42.5, AMC 2023 87.5 against 90.3), while Minerva, OlympiadBench and MATH-500 are within about a point. It is also the domain that shuffling hurts most and the one where the dense term helps earliest (62.6 against 55.1 for the sampled-token term after 100 steps). Its responses are the longest, so the per-token signal must be credited over many more positions, and thus the student's gap to the expert closes the slowest and last (see \Cref{fig:domains}).

\section{Conclusion}
\label{sec:conclusion}
We introduced \methodfull{} (\method), which consolidates RL experts by replaying their training logs through the student in the experts' order: the policy gradient is recomputed on the logged trajectories, and the final experts' distributions are distilled at the logged prefixes. Therefore, \method{} samples nothing and calls no environment or reward model, and the replay stays quasi-on-policy throughout the consolidation process. On three long-horizon agentic domains, \method{} yields the best student model, surpassing MOPD from the same experts at a third of MOPD's cost per step. Our analysis shows that the replayed trajectories carry most of the gain, and that both the policy gradient and the distillation term add to it. More importantly, the order of the logs matters more than their amount or rewards: expert training logs are a curriculum, not a data mixture. As RL experts multiply across domains and their horizons grow longer, we hope \method{} offers a simple and scalable path toward generalist long-horizon agents, in which the experience that experts have already paid for is replayed rather than resampled.

\bibliography{reference}
\bibliographystyle{icml2026}

\appendix
\crefalias{section}{appendix}   
\clearpage
\section{Implementation}
\label{app:impl}

\begin{table*}[t]
\centering
\caption{Settings. Online GRPO uses the experts' column with $3\times24\times8$ samples per step and the three environments; MOPD likewise, with the final experts resident in the trainer.}
\label{tab:settings}
\small
\adjustbox{max width=\textwidth}{%
\begin{tabular}{@{}lll@{}}
\toprule
\textbf{Setting} & \textbf{Experts (per domain)} & \textbf{Replay variants} \\
\midrule
initialization & $\piref$ (SFT cold start of Qwen3-4B) & $\piref$ \\
prompts $\times$ samples per step & $64\times8$ & $3\times64\times8$ (the experts' step-$k$ batches) \\
steps & 300 & 300 \\
learning rate & $10^{-6}$, constant & $10^{-6}$, constant \\
optimizer & Adam $(0.9, 0.98)$, weight decay 0.1 & Adam $(0.9, 0.98)$, weight decay 0.1 \\
clip $(\epsilon_{\mathrm{lo}},\epsilon_{\mathrm{hi}})$ & $(0.2,\,0.28)$ & $(0.2,\,0.28)$ (GRPO variants) \\
KL penalty to $\piref$, entropy bonus & none & none \\
response / context length & 16k (reasoning), 9k (search, SQL) & as logged \\
sampling & SGLang, temperature 1, asynchronous & none \\
distillation target & -- & the final experts (after 300 steps) \\
dense term & -- & $k=256$, reverse KL, $\beta=1$ \\
parallelism & TP 2, rollout engines on the same node & TP 2, trainer on all 8 GPUs \\
\bottomrule
\end{tabular}}
\end{table*}

The replay is implemented on slime, with the settings of \Cref{tab:settings}. The experts' rollout dumps are merged into per-step files and loaded in place of the rollout engine, and the experts are loaded as extra Megatron weight sets in the trainer, each running one forward pass per step on its domain's samples.

The baselines share this code path. Online GRPO applies the experts' recipe to the three domains at once, with 24 prompts $\times$ 8 samples per domain and step. Online PPO replaces the group normalization by a learned critic: the advantage is the return minus the value ($\gamma=\lambda=1$), whitened over the batch, and the critic takes two optimizer steps per actor step at $5\times10^{-6}$ each, half the $10^{-5}$ of the PPO experts' single update. MOPD runs the same rollout with the three final experts in the trainer, each scoring its domain's samples with the sampled-token reward at coefficient 1 and no task reward; MOPD + GRPO adds the group-normalized task reward. RAFT samples like online GRPO and trains on every correct sample with the supervised loss, as in~\citet{xiong2025minimalist}; prompts with no correct sample contribute nothing. Its loss is averaged over the kept tokens of the step, not per sample. The uniform soup averages the three expert deltas from $\piref$ (task arithmetic at $\lambda=1/3$); TIES keeps each delta's largest 20\% of entries, resolves sign conflicts by majority, and adds the result at $\lambda=1$. The PPO experts are single-domain PPO runs with the GRPO experts' rollout, $\gamma=\lambda=1$, batch-whitened advantages, and one critic update per step at learning rate $10^{-5}$.

\paragraph{Where each term enters.}
The PG term is the clipped surrogate with the group-normalized reward as the advantage: the reward is centered within the prompt's group and divided by the group's standard deviation plus $10^{-6}$, with no batch whitening. The sampled-token term is added to that advantage, $A_t \leftarrow A_t - \beta\,(\log\pis(y_t\mid s_t)-\log\piT(y_t\mid s_t))$, and so passes through the ratio and the clip. The dense term never touches the advantage: it is computed inside the policy loss and added to it, $\mathcal{L}=\mathcal{L}_{\mathrm{PG}}+\overline{\ell_{\mathrm{KL}}}$, without ratio or clip, and the sampled-token term is off whenever the dense term is on. SFT and offline RAFT replace the surrogate by the supervised loss.

\paragraph{The dense KL.}
The vocabulary-level KL of \Cref{eq:kl} is computed over the final expert's top-$k$ tokens. At every logged prefix $s_t$ with $t\in\M$, the expert's forward pass yields its $k$ most probable next tokens $\{v_1,\dots,v_k\}$ and their probabilities, and the student's probabilities at the same tokens are read from its own forward pass. Both distributions are coarsened to $k+1$ atoms, i.e., the $k$ tokens and a tail atom with the remaining mass, and the per-token term is
\begin{equation}
  \ell_{\mathrm{KL}}(s_t)\;=\;\beta\,\KL\!\big(\tilde p_\theta(\cdot\mid s_t)\,\big\|\,
  \tilde p_T(\cdot\mid s_t)\big),
  \label{eq:dense}
\end{equation}
where $\tilde p(v_j)=p(v_j)$ for $j\le k$ and $\tilde p(\mathrm{tail})=1-\sum_{j\le k}p(v_j)$. The coarsened KL lower-bounds the full-vocabulary KL and equals it at $k=V$. We use $k=256$ and $\beta=1$ unless stated otherwise. In the replay, the logged token is among its expert's top-256 at 99.9\% of the positions, so the coarsening leaves almost no sampled token in the tail. Kimi K3 reports no gain from top-$k$ objectives in on-policy distillation~\citep{team2026kimi}. There the sampled token comes from the student, so the sampled-token estimate of the reverse KL is unbiased; in the replay the token comes from the logging policy, and the dense term is better by 0.4 to 0.5 (\Cref{tab:variants}). It does not narrow the student's samples either (pass@$k$, \Cref{app:passk}).

In the implementation, the experts' forward pass runs without gradient and stores, at every response position, the ids and log-probabilities of the $k$ most probable tokens (about 2\,GB per rank per step at $k=256$), merged across the tensor-parallel vocabulary shards. The student's log-probabilities at those ids come out of the policy loss's own vocabulary-parallel log-probability function, so the term adds no full-vocabulary tensor. Each side's tail atom is $\log\big(1-\sum_{j\le k}p(v_j)\big)$, with the sum clamped at $1-10^{-6}$. The reverse, forward and Jensen--Shannon divergences are computed in closed form from the two $(k+1)$-atom vectors, with gradients through the student only. A tensor-parallel self-check recomputes the experts' atoms on one shard and logs the discrepancy (about $10^{-5}$).

\paragraph{Reduction and batch.}
Every per-token term is averaged over the model tokens of its sample, and the per-sample means are averaged over all samples of the step; a sample taken out of the loss, as in offline RAFT, contributes zero. Online RAFT is the exception: its loss is averaged over all kept tokens of the step (above). The 1536 replayed samples form one optimizer step, split into dynamic micro-batches of at most 16.4k tokens per GPU, with tensor parallelism 2 on eight GPUs and full activation recomputation.

\paragraph{Optimizer.}
Every run, the experts included, uses Adam with $(\beta_1,\beta_2)=(0.9,0.98)$, a constant learning rate of $10^{-6}$, weight decay 0.1, clip range $(0.2, 0.28)$, no entropy bonus and no KL penalty to $\piref$. The importance ratio is taken to the log-probabilities recorded by the sampling engine: the logs for the replay, and the asynchronous engine, one step behind, for the online runs.

\paragraph{Initialization.}
$\piref$ is Qwen3-4B (the Instruct-2507 release) after supervised fine-tuning on the 2000 traces of ReTool's SFT set~\citep{feng2026retool}, converted to the chat format with tool calls: three epochs at batch size 64 (93 steps), with Adam $(0.9, 0.95)$, weight decay 0.1 and a learning rate of $5\times10^{-5}$ that warms up over the first 10\% of the steps and decays by a cosine to $10^{-6}$; the loss is averaged over tokens. The traces are reasoning problems solved with the Python tool: $\piref$ has seen no search or SQL trace, and it starts low in these two domains (\Cref{tab:main}). Every model, i.e., the student, the experts and the baselines, starts from it.

\paragraph{Environments and training data.}
Every environment scores a trajectory with $R=+1$ if its final answer is correct and $-1$ otherwise. In search and SQL the grader is the evaluation's (\Cref{app:eval}); in reasoning, the training reward requires the content of the final boxed answer to equal the reference as a string (the strict grader of DAPO), which is stricter than the equivalence check of the evaluation. \emph{Reasoning}: the prompts are 6400 problems of DAPO-Math-17k, and the model runs up to 16 code turns in the Python sandbox within 16k generated tokens. \emph{Search}: 6500 questions of the NQ and HotpotQA training sets of Search-R1~\citep{jin2025search,yue2026dr}; a query returns the top 3 passages of the E5 retriever over the 2018 Wikipedia dump, with up to 8 search turns of at most 4k tokens each within a 9k context. \emph{SQL}: 6400 questions of the Spider training set over 140 databases, sampled after executing the gold queries (those that fail or return nothing, and repeated questions, are dropped); the prompt holds the database's \texttt{CREATE TABLE} statements with the first three rows of each table, and the model may run up to 6 read-only queries within 8 turns, under the token limits of search, each result shown up to 20 rows. BIRD is not trained on. Each expert draws 64 prompts per step from its shuffled set, so its 300 steps are about three passes over it; the online baselines, at 24 prompts per domain and step, pass over each set about once.

\section{Evaluation Protocol}
\label{app:eval}

\Cref{tab:protocol} lists the sixteen sets with their number of problems and of samples per problem. Every checkpoint, the experts and $\piref$ included, is served with SGLang and sampled at temperature 1 and top-$p$ 1 in the environment it was trained in: the Python sandbox with up to 16 code turns and 16k generated tokens for reasoning; the E5 retriever over Wikipedia with 3 passages per query, up to 8 search turns and 8k generated tokens (4k per turn) within a 9k context for search; and the read-only SQLite tool with up to 8 turns and the same token limits for SQL. On the six large search sets, 2500 questions are drawn once with a fixed seed, so every checkpoint answers the same questions; Bamboogle is used whole. BIRD dev in the tables is the 500-question mini-dev release, on 11 databases, with the benchmark's evidence given as a hint in the prompt. BIRD is used for evaluation only: the SQL prompts of training are Spider's (\Cref{app:impl}), and the databases of the three SQL evaluation sets are disjoint from the training ones. The checkpoint evaluated is the one after 300 steps, and also those after 100 and 200 steps in \Cref{tab:variants}.

\begin{table}[t]
\centering
\caption{The evaluation sets: problems, samples per problem ($N$) and traces per checkpoint.}
\label{tab:protocol}
\small
\adjustbox{max width=\columnwidth}{%
\begin{tabular}{@{}llrrr@{}}
\toprule
\textbf{Domain} & \textbf{Set} & \textbf{Problems} & $\boldsymbol{N}$ & \textbf{Traces} \\
\midrule
Reasoning & AIME 2024 & 30 & 32 & 960 \\
 & AIME 2025 & 30 & 32 & 960 \\
 & AMC 2023 & 40 & 32 & 1280 \\
 & Minerva & 272 & 4 & 1088 \\
 & OlympiadBench & 675 & 4 & 2700 \\
 & MATH-500 & 500 & 4 & 2000 \\
\midrule
Search & NQ, TriviaQA, PopQA & 2500 each & 1 & 7500 \\
 & HotpotQA, 2WikiMQA, MuSiQue & 2500 each & 1 & 7500 \\
 & Bamboogle & 125 & 1 & 125 \\
\midrule
SQL & Spider dev & 1034 & 4 & 4136 \\
 & Spider test & 2147 & 4 & 8588 \\
 & BIRD mini-dev & 500 & 4 & 2000 \\
\bottomrule
\end{tabular}}
\end{table}

\paragraph{Scores.}
A problem's score is the fraction of its $N$ samples that are correct, and a set's score is the mean over its problems, every problem with the same weight: avg@$N$, the expected pass@1 of a single sample. The domain average is the unweighted mean over the sets of the domain (six, seven and three), and Overall is the mean of the three domain averages. Treating the traces of \Cref{tab:protocol} as independent, the binomial standard error of Overall is about 0.3. All numbers are single runs; a second run of \method{} on the same replayed batches reaches 58.5 against 58.6 (by set in \Cref{app:sets}).

\paragraph{Graders.}
In reasoning, the final boxed answer is compared with the reference by a symbolic and numeric equivalence check (the grader of OAT, built on Math-Verify); a response without a boxed answer is wrong. In search, the answer in the final \texttt{<answer>} tag, lower-cased and with punctuation and articles removed, must equal one of the gold answers: the normalized exact match of Search-R1, which is also the training reward, with no partial credit for format or retrieval. In SQL, the final query (the last \texttt{<answer>} block, or else the last SQL code block of the final turn) and the gold query are executed on the database and their result sets compared, as ordered lists when the gold query has \texttt{ORDER BY} and as multisets otherwise, which is Spider's execution accuracy; a query that fails, or a response without one, is wrong.

\section{Additional Results}
\label{app:sets}
\Cref{tab:math,tab:search,tab:sql} list every run of \Cref{tab:main,tab:variants,tab:order,tab:ablation} on all sixteen sets after 300 steps, under the protocol of \Cref{app:eval}, and the second run of \method{} by set, for comparison with the first; the replayed batches of the two runs are the same. They also list the sampled-token KL with the PG term at $\beta=0.3$, which is not in \Cref{tab:ablation}.

\begin{table}[t]
\centering
\caption{Generation of the online runs by phase, as medians over the steps of each phase: the share of search trajectories cut off at the turn limit, the mean length of the reasoning trajectories in tokens, the tokens of the step's 576 trajectories, the speed of the step's longest trajectory in tokens per second, the rollout time and the step time in seconds.}
\label{tab:steptime}
\small
\adjustbox{max width=\columnwidth}{%
\begin{tabular}{@{}lcccccc@{}}
\toprule
\textbf{Steps} & \makecell{\textbf{Search}\\\textbf{at Limit}} & \makecell{\textbf{Reasoning}\\\textbf{Length}} & \makecell{\textbf{Tokens}\\\textbf{/ Step}} & \makecell{\textbf{Longest}\\\textbf{(tok/s)}} & \makecell{\textbf{Rollout}\\\textbf{(s)}} & \makecell{\textbf{Step}\\\textbf{(s)}} \\
\midrule
\multicolumn{7}{@{}l}{\textit{Online GRPO}} \\
\quad 0--175 & 3\% & 3.1k & 1.6M & 102 & 154 & 153 \\
\quad 225--300 & 86\% & 6.8k & 2.5M & 36 & 446 & 445 \\
\midrule
\multicolumn{7}{@{}l}{\textit{Online PPO}} \\
\quad 0--175 & 2\% & 3.0k & 1.5M & 89 & 171 & 183 \\
\quad 225--250 & 85\% & 8.8k & 2.9M & 31 & 525 & 534 \\
\quad 275--300 & 82\% & 4.8k & 2.1M & 35 & 456 & 469 \\
\midrule
\multicolumn{7}{@{}l}{\textit{MOPD + GRPO}} \\
\quad 0--50 & 1\% & 4.4k & 1.8M & 89 & 177 & 199 \\
\quad 60--175 & 80\% & 7.4k & 2.6M & 38 & 419 & 427 \\
\quad 250--300 & 14\% & 7.7k & 2.5M & 77 & 209 & 270 \\
\bottomrule
\end{tabular}}
\end{table}

\begin{table*}[t]
\centering
\caption{Reasoning after 300 steps: accuracy averaged over 32 samples per problem on AIME 2024, AIME 2025 and AMC 2023, and over 4 elsewhere.}
\label{tab:math}
\small
\adjustbox{max width=\textwidth}{%
\begin{tabular}{@{}lccccccc@{}}
\toprule
\textbf{Method} & \textbf{AIME24} & \textbf{AIME25} & \textbf{AMC23} & \textbf{Minerva} & \textbf{Olympiad} & \textbf{MATH-500} & \textbf{Average} \\
\midrule
\multicolumn{8}{@{}l}{\textit{Baselines}} \\
\quad SFT initialization & 26.9 & 20.3 & 66.6 & 38.3 & 50.4 & 81.3 & 47.3 \\
\quad merging, uniform soup & 39.0 & 31.4 & 78.8 & 42.9 & 59.4 & 87.8 & 56.6 \\
\quad merging, TIES & 47.0 & 38.0 & 84.2 & 40.9 & 62.2 & 87.9 & 60.0 \\
\quad RAFT & 41.0 & 32.4 & 78.9 & 42.4 & 60.1 & 88.1 & 57.2 \\
\quad online GRPO & 40.3 & 32.8 & 76.0 & 30.9 & 47.1 & 74.5 & 50.3 \\
\quad online PPO & 39.5 & 31.9 & 77.7 & 33.6 & 54.1 & 82.0 & 53.1 \\
\quad MOPD & 48.1 & 42.4 & 87.8 & 45.0 & 67.3 & 92.3 & 63.8 \\
\quad MOPD + GRPO & 48.6 & 40.7 & 87.6 & 44.3 & 66.1 & 93.0 & 63.4 \\
\midrule
\multicolumn{8}{@{}l}{\textit{Without the KL term (\Cref{tab:order})}} \\
\quad SFT on the logs & 47.8 & 40.4 & 85.9 & 44.7 & 65.0 & 91.5 & 62.6 \\
\quad SFT on the logs, steps shuffled & 41.2 & 34.7 & 80.9 & 40.6 & 60.3 & 88.3 & 57.7 \\
\quad RAFT on the logs & 48.6 & 41.7 & 85.4 & 43.5 & 65.7 & 92.0 & 62.8 \\
\quad RAFT on the logs, steps shuffled & 44.6 & 35.0 & 82.6 & 41.4 & 62.7 & 90.0 & 59.4 \\
\quad PG only (offline GRPO) & 48.6 & 40.1 & 86.2 & 43.5 & 65.3 & 91.8 & 62.6 \\
\quad PG only, steps shuffled & 42.3 & 35.8 & 81.2 & 41.9 & 60.9 & 88.0 & 58.4 \\
\midrule
\multicolumn{8}{@{}l}{\textit{Variants of \method{} (\Cref{tab:variants})}} \\
\quad sampled-token KL only & 49.6 & 40.4 & 87.7 & 43.6 & 65.1 & 91.0 & 62.9 \\
\quad PG + sampled-token KL & 49.6 & 41.1 & 86.9 & 45.0 & 66.0 & 92.6 & 63.5 \\
\quad dense reverse KL only & 50.3 & 40.9 & 86.8 & 43.4 & 66.9 & 92.9 & 63.5 \\
\quad PG + dense reverse KL (\method) & 50.4 & 40.7 & 87.5 & 44.0 & 67.0 & 92.8 & 63.7 \\
\quad \method, second run & 50.5 & 41.6 & 88.0 & 44.5 & 66.4 & 92.0 & -- \\
\quad dense forward KL only & 48.6 & 41.9 & 87.3 & 44.2 & 65.9 & 92.5 & 63.4 \\
\quad PG + dense forward KL & 48.9 & 40.7 & 87.0 & 43.0 & 64.6 & 91.7 & 62.7 \\
\midrule
\multicolumn{8}{@{}l}{\textit{Ablations (\Cref{tab:ablation})}} \\
\quad \method, $\beta=3$ & 50.2 & 42.0 & 88.2 & 43.8 & 66.1 & 91.9 & 63.7 \\
\quad \method, $\beta=0.3$ & 48.5 & 40.5 & 87.2 & 43.3 & 65.0 & 92.2 & 62.8 \\
\quad \method, $\beta=0.1$ & 48.4 & 41.1 & 86.3 & 43.9 & 64.7 & 90.6 & 62.5 \\
\quad \method, 576 / step & 49.7 & 41.0 & 87.3 & 45.1 & 66.4 & 92.3 & 63.6 \\
\quad \method, steps shuffled & 49.3 & 39.7 & 89.0 & 43.2 & 66.0 & 92.2 & 63.2 \\
\quad PG + sampled-token KL, $\beta=0.3$ & 49.1 & 40.2 & 86.3 & 42.1 & 64.5 & 90.2 & 62.1 \\
\quad PG + sampled-token KL, 576 / step & 49.6 & 42.9 & 86.9 & 41.0 & 65.1 & 91.1 & 62.8 \\
\quad PG + sampled-token KL, steps shuffled & 47.5 & 38.1 & 86.3 & 42.8 & 64.0 & 90.0 & 61.4 \\
\midrule
\multicolumn{8}{@{}l}{\textit{Single-domain references}} \\
\quad GRPO expert of the domain & 53.8 & 42.5 & 90.3 & 45.1 & 66.3 & 93.4 & 65.2 \\
\quad PPO expert of the domain & 43.8 & 44.6 & 89.1 & 44.8 & 66.5 & 93.0 & 63.6 \\
\bottomrule
\end{tabular}
}
\end{table*}

\begin{table*}[t]
\centering
\caption{Search after 300 steps: exact match, one sample, up to 2500 questions per set (Bamboogle 125).}
\label{tab:search}
\small
\adjustbox{max width=\textwidth}{%
\begin{tabular}{@{}lcccccccc@{}}
\toprule
\textbf{Method} & \textbf{NQ} & \textbf{TriviaQA} & \textbf{PopQA} & \textbf{HotpotQA} & \textbf{2WikiMQA} & \textbf{MuSiQue} & \textbf{Bamboogle} & \textbf{Average} \\
\midrule
\multicolumn{9}{@{}l}{\textit{Baselines}} \\
\quad SFT initialization & 17.0 & 44.8 & 19.0 & 19.2 & 26.2 & 5.6 & 24.8 & 22.4 \\
\quad merging, uniform soup & 23.8 & 52.4 & 27.5 & 25.8 & 31.6 & 9.6 & 35.2 & 29.4 \\
\quad merging, TIES & 38.6 & 66.0 & 41.8 & 40.6 & 44.4 & 18.6 & 48.0 & 42.6 \\
\quad RAFT & 29.3 & 59.6 & 32.0 & 31.7 & 34.9 & 11.0 & 40.0 & 34.1 \\
\quad online GRPO & 35.7 & 64.6 & 38.5 & 37.9 & 41.0 & 16.3 & 44.0 & 39.7 \\
\quad online PPO & 32.7 & 62.4 & 36.3 & 36.0 & 39.8 & 16.2 & 44.8 & 38.3 \\
\quad MOPD & 37.0 & 62.6 & 39.1 & 40.6 & 42.3 & 17.7 & 44.8 & 40.6 \\
\quad MOPD + GRPO & 38.4 & 64.6 & 38.7 & 40.0 & 43.5 & 18.3 & 46.4 & 41.4 \\
\midrule
\multicolumn{9}{@{}l}{\textit{Without the KL term (\Cref{tab:order})}} \\
\quad SFT on the logs & 38.5 & 63.4 & 39.5 & 40.5 & 43.0 & 18.0 & 40.8 & 40.5 \\
\quad SFT on the logs, steps shuffled & 34.5 & 61.3 & 37.4 & 37.8 & 41.0 & 15.3 & 40.8 & 38.3 \\
\quad RAFT on the logs & 38.1 & 63.0 & 40.0 & 40.9 & 41.8 & 18.4 & 44.8 & 41.0 \\
\quad RAFT on the logs, steps shuffled & 36.1 & 62.4 & 38.0 & 37.4 & 40.0 & 15.8 & 41.6 & 38.8 \\
\quad PG only (offline GRPO) & 39.3 & 64.2 & 39.9 & 41.2 & 43.5 & 18.0 & 40.8 & 41.0 \\
\quad PG only, steps shuffled & 35.2 & 62.6 & 36.8 & 39.8 & 41.2 & 15.9 & 45.6 & 39.6 \\
\midrule
\multicolumn{9}{@{}l}{\textit{Variants of \method{} (\Cref{tab:variants})}} \\
\quad sampled-token KL only & 38.0 & 63.2 & 39.0 & 40.8 & 43.6 & 18.0 & 43.2 & 40.8 \\
\quad PG + sampled-token KL & 38.5 & 63.0 & 40.3 & 40.4 & 43.9 & 18.6 & 45.6 & 41.5 \\
\quad dense reverse KL only & 39.7 & 63.9 & 40.2 & 41.0 & 43.2 & 18.2 & 44.8 & 41.6 \\
\quad PG + dense reverse KL (\method) & 38.3 & 64.4 & 39.8 & 42.1 & 44.5 & 17.8 & 47.2 & 42.0 \\
\quad \method, second run & 38.6 & 63.6 & 40.1 & 40.4 & 43.6 & 18.6 & 48.8 & -- \\
\quad dense forward KL only & 38.8 & 64.1 & 40.7 & 41.0 & 44.5 & 19.0 & 44.0 & 41.7 \\
\quad PG + dense forward KL & 38.7 & 64.4 & 40.2 & 40.8 & 43.1 & 18.6 & 44.0 & 41.4 \\
\midrule
\multicolumn{9}{@{}l}{\textit{Ablations (\Cref{tab:ablation})}} \\
\quad \method, $\beta=3$ & 39.5 & 63.8 & 40.0 & 40.5 & 43.9 & 18.6 & 42.4 & 41.2 \\
\quad \method, $\beta=0.3$ & 39.2 & 64.2 & 39.7 & 41.0 & 42.2 & 18.9 & 46.4 & 41.6 \\
\quad \method, $\beta=0.1$ & 38.7 & 63.8 & 40.6 & 40.8 & 44.0 & 18.0 & 45.6 & 41.7 \\
\quad \method, 576 / step & 39.2 & 64.1 & 40.2 & 41.8 & 42.8 & 18.2 & 44.8 & 41.6 \\
\quad \method, steps shuffled & 37.8 & 63.6 & 40.4 & 40.6 & 43.6 & 17.3 & 46.4 & 41.4 \\
\quad PG + sampled-token KL, $\beta=0.3$ & 37.9 & 63.5 & 40.3 & 41.3 & 43.5 & 18.1 & 46.4 & 41.6 \\
\quad PG + sampled-token KL, 576 / step & 37.9 & 63.3 & 40.2 & 40.8 & 43.0 & 18.7 & 41.6 & 40.8 \\
\quad PG + sampled-token KL, steps shuffled & 37.9 & 62.7 & 39.7 & 40.8 & 43.4 & 17.7 & 44.8 & 41.0 \\
\midrule
\multicolumn{9}{@{}l}{\textit{Single-domain references}} \\
\quad GRPO expert of the domain & 39.6 & 65.1 & 40.6 & 42.3 & 44.2 & 18.7 & 50.4 & 43.0 \\
\quad PPO expert of the domain & 36.3 & 62.4 & 39.0 & 40.2 & 41.4 & 16.8 & 42.4 & 39.8 \\
\bottomrule
\end{tabular}
}
\end{table*}

\begin{table*}[t]
\centering
\caption{SQL after 300 steps: execution accuracy over 4 samples.}
\label{tab:sql}
\small
\adjustbox{max width=\textwidth}{%
\begin{tabular}{@{}lcccc@{}}
\toprule
\textbf{Method} & \textbf{Spider dev} & \textbf{Spider test} & \textbf{BIRD dev} & \textbf{Average} \\
\midrule
\multicolumn{5}{@{}l}{\textit{Baselines}} \\
\quad SFT initialization & 17.3 & 20.0 & 7.5 & 14.9 \\
\quad merging, uniform soup & 61.7 & 63.9 & 31.8 & 52.5 \\
\quad merging, TIES & 76.1 & 77.8 & 49.1 & 67.7 \\
\quad RAFT & 72.3 & 73.3 & 43.9 & 63.2 \\
\quad online GRPO & 76.0 & 77.2 & 49.1 & 67.4 \\
\quad online PPO & 75.1 & 76.6 & 47.3 & 66.3 \\
\quad MOPD & 80.1 & 80.1 & 49.4 & 69.9 \\
\quad MOPD + GRPO & 79.6 & 79.9 & 49.6 & 69.7 \\
\midrule
\multicolumn{5}{@{}l}{\textit{Without the KL term (\Cref{tab:order})}} \\
\quad SFT on the logs & 79.8 & 79.8 & 48.3 & 69.3 \\
\quad SFT on the logs, steps shuffled & 74.7 & 76.5 & 44.2 & 65.1 \\
\quad RAFT on the logs & 79.2 & 79.8 & 48.2 & 69.1 \\
\quad RAFT on the logs, steps shuffled & 76.8 & 77.8 & 46.2 & 66.9 \\
\quad PG only (offline GRPO) & 79.5 & 79.1 & 50.1 & 69.6 \\
\quad PG only, steps shuffled & 74.8 & 76.3 & 46.2 & 65.8 \\
\midrule
\multicolumn{5}{@{}l}{\textit{Variants of \method{} (\Cref{tab:variants})}} \\
\quad sampled-token KL only & 79.8 & 79.3 & 49.5 & 69.5 \\
\quad PG + sampled-token KL & 80.6 & 80.1 & 48.4 & 69.7 \\
\quad dense reverse KL only & 80.0 & 79.9 & 48.5 & 69.5 \\
\quad PG + dense reverse KL (\method) & 80.7 & 79.8 & 49.9 & 70.1 \\
\quad \method, second run & 80.2 & 80.0 & 48.5 & -- \\
\quad dense forward KL only & 79.9 & 80.1 & 49.7 & 69.9 \\
\quad PG + dense forward KL & 79.9 & 79.9 & 48.6 & 69.5 \\
\midrule
\multicolumn{5}{@{}l}{\textit{Ablations (\Cref{tab:ablation})}} \\
\quad \method, $\beta=3$ & 80.3 & 79.6 & 48.6 & 69.5 \\
\quad \method, $\beta=0.3$ & 80.9 & 80.0 & 49.0 & 70.0 \\
\quad \method, $\beta=0.1$ & 79.8 & 79.8 & 49.0 & 69.5 \\
\quad \method, 576 / step & 80.1 & 79.8 & 50.4 & 70.1 \\
\quad \method, steps shuffled & 79.3 & 80.0 & 48.9 & 69.4 \\
\quad PG + sampled-token KL, $\beta=0.3$ & 79.5 & 79.9 & 50.1 & 69.8 \\
\quad PG + sampled-token KL, 576 / step & 79.8 & 79.7 & 49.1 & 69.5 \\
\quad PG + sampled-token KL, steps shuffled & 80.0 & 79.7 & 48.2 & 69.3 \\
\midrule
\multicolumn{5}{@{}l}{\textit{Single-domain references}} \\
\quad GRPO expert of the domain & 80.3 & 80.0 & 48.6 & 69.6 \\
\quad PPO expert of the domain & 79.7 & 79.1 & 47.8 & 68.8 \\
\bottomrule
\end{tabular}
}
\end{table*}

\section{Pass@$k$}
\label{app:passk}
\Cref{tab:passk} reports pass@$k$, the probability that at least one of $k$ samples solves a problem, estimated without bias from its $N$ samples as $1-\binom{N-c}{k}/\binom{N}{k}$ when $c$ of them are correct~\citep{chen2021evaluating}, and averaged over problems and sets as the scores of \Cref{app:eval}. On reasoning we use the three sets with 32 samples per problem (AIME 2024, AIME 2025 and AMC 2023), on SQL all three sets (4 samples); search, with one sample per question, has pass@1 only.

The students gain most at small $k$. On reasoning, \method{} leads online GRPO by 9.8 at pass@1 and by 1.4 at pass@32; on SQL, by 2.7 at pass@1, and the two are level at pass@4. The dense KL does not narrow the student: at pass@32, \method{} is level with MOPD and above the sampled-token KL, SFT on the logs, online GRPO and $\piref$. The mass-covering forward KL leads from pass@8 on reasoning (87.8 against 86.9 at pass@32), but not on SQL.

\begin{table*}[t]
\centering
\caption{Pass@$k$ in percent after 300 steps: on reasoning the mean over AIME 2024, AIME 2025 and AMC 2023 (32 samples per problem), on SQL the mean over the three sets (4 samples). Pass@1 is the score of the main tables, on reasoning averaged over three sets rather than six. We mark the best of each column in bold and underline the second best.}
\label{tab:passk}
\small
\adjustbox{max width=\textwidth}{%
\begin{tabular}{@{}lcccccccc@{}}
\toprule
 & \multicolumn{5}{c}{\textbf{Reasoning}} & \multicolumn{3}{c}{\textbf{SQL}} \\
\cmidrule(lr){2-6}\cmidrule(lr){7-9}
\textbf{Method} & \textbf{pass@1} & \textbf{pass@4} & \textbf{pass@8} & \textbf{pass@16} & \textbf{pass@32} & \textbf{pass@1} & \textbf{pass@2} & \textbf{pass@4} \\
\midrule
\multicolumn{9}{@{}l}{\textit{Baselines}} \\
\quad SFT initialization & 37.9 & 60.8 & 68.3 & 74.8 & 81.1 & 14.9 & 25.1 & 38.0 \\
\quad online GRPO & 49.7 & 71.6 & 78.0 & 82.9 & 85.6 & 67.4 & 74.8 & \textbf{79.7} \\
\quad MOPD & \underline{59.4} & 75.2 & 79.8 & 83.2 & \underline{86.9} & \underline{69.9} & 75.5 & \underline{79.3} \\
\midrule
\multicolumn{9}{@{}l}{\textit{Replay of the experts' logs}} \\
\quad SFT on the logs & 58.1 & 73.8 & 78.6 & 82.2 & 84.4 & 69.3 & 75.1 & 78.9 \\
\quad PG + sampled-token KL & 59.2 & 75.2 & 79.5 & 82.4 & 84.7 & 69.7 & \underline{75.6} & \textbf{79.7} \\
\quad PG + dense forward KL & 58.9 & \underline{75.4} & \textbf{80.8} & \textbf{84.9} & \textbf{87.8} & 69.5 & 75.4 & 79.2 \\
\quad PG + dense reverse KL (\method) & \textbf{59.5} & \textbf{75.6} & \underline{80.2} & \underline{83.6} & \underline{86.9} & \textbf{70.1} & \textbf{75.8} & \textbf{79.7} \\
\bottomrule
\end{tabular}}
\end{table*}

\section{Step Time During Training}
\label{app:steptime}

\begin{figure}[t]
\centering
\includegraphics[width=\columnwidth]{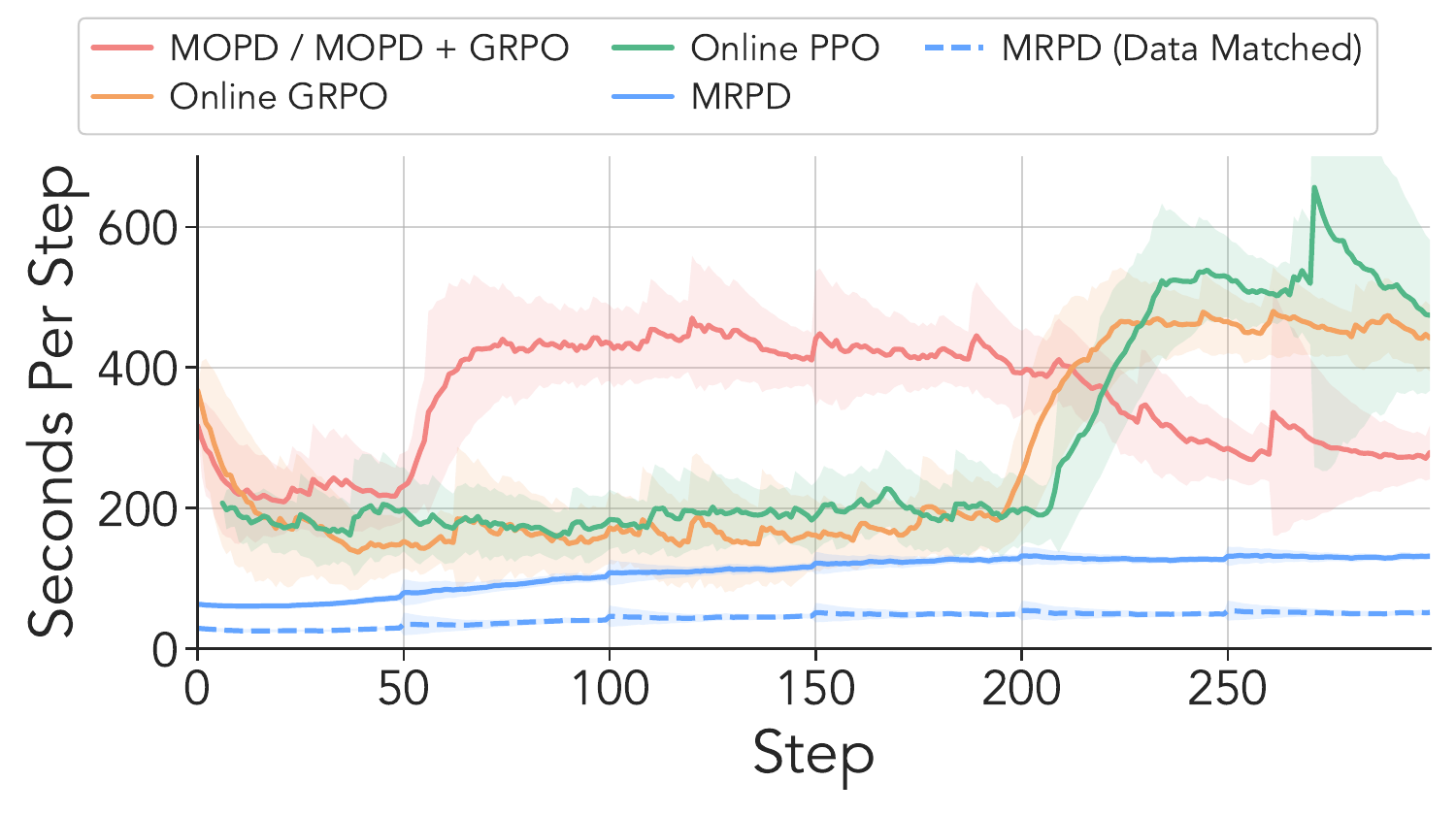}
\caption{Seconds per step on eight H100s over the 300 steps (smoothed); \Cref{fig:cost} shows their totals and their maxima. The online methods sample at every step and wait for the longest trajectory, which is generated more slowly once the search trajectories exhaust their turns and the reasoning responses lengthen (\Cref{tab:steptime}); MOPD stands for its run with the task reward and online GRPO for a run with the rewards normalized by the group mean, with the same rollouts. \method{} replays the logs, with their full 1536 samples per step or data-matched to the online runs (576 per step, dashed).}
\label{fig:steptime}
\end{figure}

\Cref{fig:steptime} shows the seconds per step behind the totals and the maxima of \Cref{fig:cost}.

\paragraph{The runs.}
All five runs use eight H100s. MOPD is timed on its run with the task reward (MOPD + GRPO), which samples, calls the tools and runs the experts exactly as MOPD does; its history is two runs, the second resumed from the checkpoint of step 150. Online GRPO is timed on a run that normalizes the group's rewards by their mean rather than their standard deviation, with the same rollouts. MOPD itself and the GRPO run of \Cref{tab:main} were trained on A100s, at 690s and 460s per step, and are left out: their seconds are not comparable. \method{} is timed on its second run and the data-matched replay on a re-run at 576 samples per step. Both use the final implementation, which does the same computation without three kinds of waste: each expert forwards only its own domain's samples, the student's separate forward pass for the sampled-token term is skipped, and the next step's log is read in the background. The first run, whose accuracy the tables report, took 179s per step with the earlier implementation. Every 50th step also writes a checkpoint (50 to 60s), which the totals include.

\paragraph{What a step consists of.}
Training is asynchronous: the rollouts of the next step are generated while the current update runs, so a step of an online method lasts as long as the longer of the two, which is almost always the generation. A step of MOPD + GRPO takes 353s on average and its rollouts 336s. The trainer's side takes about 250s (the update 124s, the three experts' forward passes on every sampled token 96s, the student's forward pass for the sampled-token term 31s) and then waits for the rollouts. For online GRPO and PPO, the rollouts take 264 and 277s of the 266 and 289s. A step of \method{} has no rollout: it is the update (79s), the experts' forward passes on their own domain's logged batch (24s) and the reading of the next step's log (6s), 110s in all. At 576 samples per step it takes 44s (the update 30s, the experts' forward 9s).

\paragraph{Why the online runs slow down.}
A step's generation ends with its longest trajectory, and throughout training this is a reasoning response at or near the 16k-token limit: before the slow phases, the longest trajectory of a step has a median of 15.1k to 16.0k tokens, and in the slow phases it is at the limit on every step. The length of the longest trajectory therefore does not explain the slowdown; the speed at which it is generated does. \Cref{tab:steptime} gives the medians by phase. In every phase, the length of the longest trajectory divided by its speed is the rollout time to within a few seconds: 16k tokens take 157s at 102 tokens per second, and 446s at 36.

Two changes of the policy lower that speed, and they arrive together, within the same 25 to 50 steps in every run. First, the search trajectories exhaust their eight turns (\Cref{fig:truncation}): 80 to 86\% of them are cut off at the turn limit, against 1 to 3\% before. Under PPO the share rises from 0.01 to 0.9 within 25 steps, and under GRPO within 50; the search expert's rose to 0.6 at step 150, so this is what RL learns on the task. Second, the reasoning responses lengthen, from 3--4k to 7--9k tokens on average, and up to 41\% of them run to the 16k limit. Both load the engines, which are shared by the 576 trajectories of a step. The tokens of a step grow from between 1.5M and 1.8M to between 2.5M and 2.9M, and a response is decoded more slowly the more sequences are generated beside it. A search trajectory at the limit also issues eight retrieval queries and eight generation calls, each on its whole context of retrieved passages so far, on GPUs that the retriever shares with the engines; the reasoning response that ends the step waits for no tool itself, but is decoded on the same GPUs. The retrieval calls slow down too: the time a trajectory waits for its tools rises from 0.1s on average to between 1 and 2.4s, and to between 60 and 90s for the slowest.

The two causes can be told apart where one changes without the other. In MOPD + GRPO, the reasoning responses stay at about 7.5k tokens from step 50 to the end, and the rollout time falls from 419 to 209s as the share of search trajectories at the limit falls from 80 to 14\%. Under PPO, the search trajectories stay at the limit from step 225 on, and the rollout time falls from 525 to 456s as the reasoning responses shorten from 8.8k to 4.8k tokens. A least-squares fit of the log rollout time on the two quantities, over the steps of each run, agrees: it gives a factor of 2.2 to 2.8 for the search trajectories going from none to 85\% at the limit, and a factor of 1.1 to 1.4 for reasoning responses longer by 4k tokens. Both slow the online runs down, the search turns more than the reasoning length. Over the last 50 steps of MOPD + GRPO the rollouts take 209s, and the step, at 270s, is bound by the trainer's side instead.

\paragraph{The replay.}
A step of \method{} follows the logged batch alone: its time follows the number of logged tokens, not the longest response. It grows from 60 to 130s as the experts' logged responses lengthen, and from 25 to 50s at 576 samples per step, without the jumps of the online runs.

\paragraph{The longest steps.}
The curves of \Cref{fig:steptime} are smoothed and hide single steps; the right panel of \Cref{fig:cost} shows the longest step of each run. For every online run it is a slow rollout: 878s for MOPD + GRPO, of which the rollout is 861s, and 626s for online GRPO, of which it is 624s. Online PPO's, at 1887s, stands apart from the rest of its run: at step 271 the longest trajectory was generated at 8.5 tokens per second, against 33 on the steps around it, and the next longest step takes 766s. The longest steps of \method{} are the ones that also write a checkpoint, up to 173s, and no other step takes more than 142s; at 576 samples per step these are 98 and 57s.

\section{Token Entropy}
\label{app:entropy}

\begin{figure}[t]
\centering
\includegraphics[width=\columnwidth]{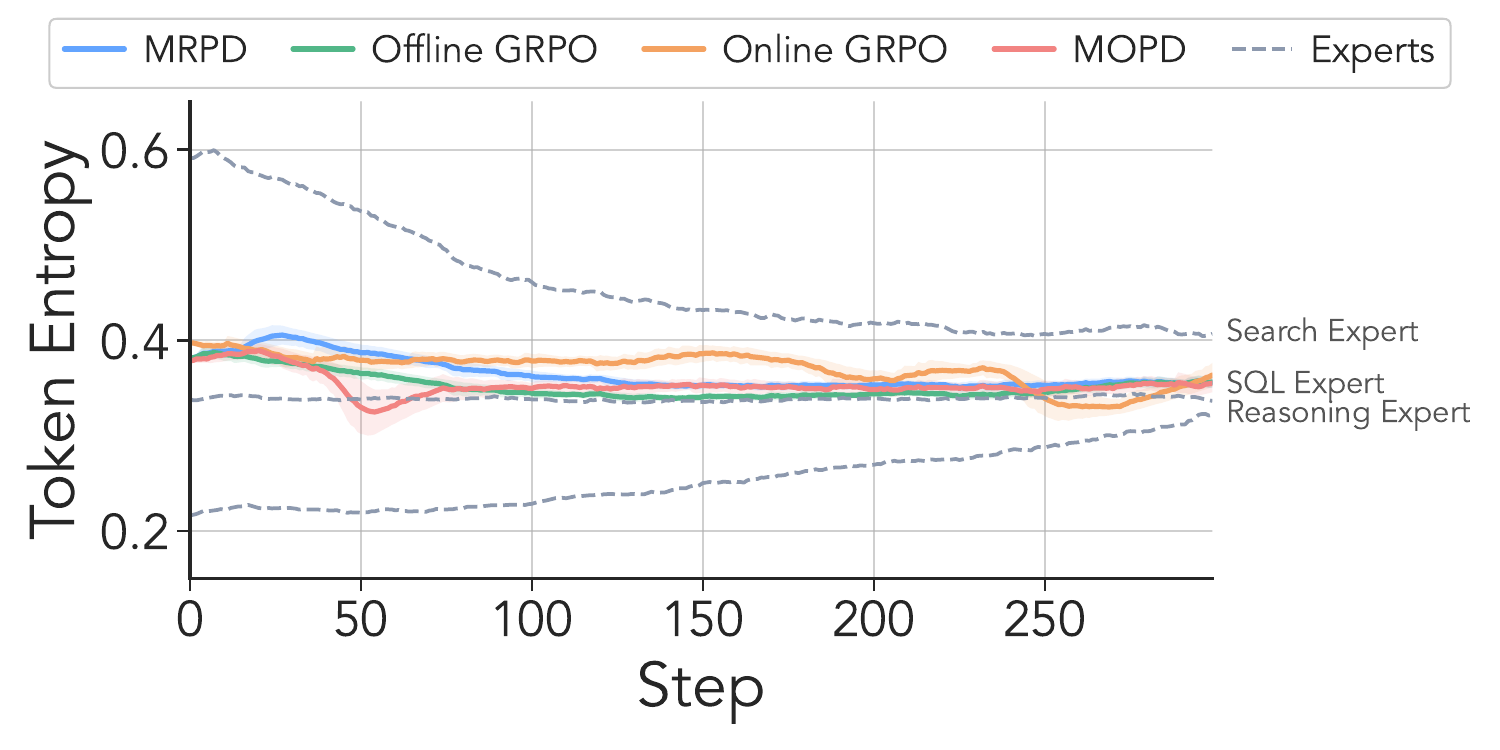}
\caption{Token entropy of the student during training: \method{}, offline GRPO on the same logs, online GRPO and MOPD. Dashed: the single-domain GRPO experts.}
\label{fig:entropy}
\end{figure}

\Cref{fig:entropy} tracks the student's token entropy. It neither collapses under the RL loss nor rises under the distillation term, which matches full next-token distributions rather than sampled tokens. The entropy of \method{} moves from 0.38 to 0.36, between the experts' 0.32 (reasoning), 0.34 (SQL) and 0.41 (search); offline GRPO ends at the same level, MOPD at 0.35 after a dip to 0.30 near step 50, and online GRPO at 0.36.

\section{Limitations}
\label{app:limitations}

\paragraph{Shared initialization.}
Every run in this paper starts from one SFT checkpoint, $\piref$: the experts are trained from it, and the student replays their logs from it. The replay relies on this. At the first step the student is the policy that sampled the logged batch, and following the experts' order keeps it close to the logging policies afterwards, which is why the importance ratio stays largely within the clip range (\Cref{sec:analysis}). We have not measured how sensitive \method{} is to this condition: neither a student initialized from another checkpoint, nor experts trained from checkpoints other than the student's, has been tried. The shuffled replay gives a partial indication of what happens when the student is not close to the logging policies: the mismatch stays high, and the student loses 0.6 points with the dense KL and 2.6 to 3.8 points without the KL term (\Cref{tab:ablation,tab:order}). We therefore expect the dense KL, which does not pass through the ratio, to matter more as the initializations move apart, but this remains to be tested.

\paragraph{Scale.}
Compute constraints limit the study to a 4B dense student with experts of the same size, three domains and 300 steps, and to a single run of each configuration (a second run of \method{} reaches 58.5 against 58.6, \Cref{app:eval}). Whether the findings carry to larger models and to mixture-of-experts models, where consolidation is used in practice, remains to be shown: that the replay stays quasi-on-policy, that it surpasses MOPD at a third of its cost per step, and that the order of the logs matters more than their amount or rewards. A replay step involves no generation and no tool call, so we expect its cost advantage to hold at scale, while the effect of model size and architecture on the mismatch is unknown.

\end{document}